\documentclass{article}

\usepackage{microtype}
\usepackage{graphicx}
\usepackage{subfigure}
\usepackage{booktabs} 
\usepackage{longtable}
\usepackage{makecell}

\usepackage{hyperref}
\usepackage{colortbl}
\usepackage{xcolor}
\usepackage{array}
\usepackage{url}
\usepackage[dvipsnames]{xcolor}
\usepackage{caption}
\usepackage{algorithmic}
\usepackage{algorithm2e}
\usepackage{enumitem}
\usepackage{multirow}

\definecolor{tticblue}{RGB}{0, 94, 184}
\definecolor{red}{RGB}{184, 0, 94}
\newcommand{\component}[1]{\textsc{#1}\xspace}   

\newcommand{\subcomponent}[1]{\text{#1}\xspace}

\newcolumntype{Z}{>{\setbox0=\hbox\bgroup}c<{\egroup}@{}}
\newcolumntype{R}{>{\raggedleft\arraybackslash}c}

\newcommand{\tableskip}{\vskip 0.15in}

\colorlet{c1}{Red!30}
\colorlet{c2}{Green!30}
\colorlet{c3}{Blue!30}
\colorlet{c4}{Yellow!30}
\colorlet{c5}{Purple!30}
\colorlet{c6}{Cyan!30}
\colorlet{c7}{Magenta!30}
\colorlet{c8}{Orange!30}
\colorlet{c9}{Turquoise!30}
\colorlet{c10}{LimeGreen!30}
\colorlet{c11}{Violet!30}
\colorlet{c12}{Brown!30}
\colorlet{c13}{Gray!30}
\colorlet{c14}{Apricot!30}
\colorlet{c15}{Aquamarine!30}
\colorlet{c16}{BlueViolet!30}
\colorlet{c17}{CarnationPink!30}
\colorlet{c18}{Emerald!30}
\colorlet{c19}{Goldenrod!30}
\colorlet{c20}{Lavender!30}
\colorlet{c21}{Peach!30}
\colorlet{c22}{Periwinkle!30}
\colorlet{c23}{RawSienna!30}
\colorlet{c24}{SeaGreen!30}
\colorlet{c25}{Thistle!30}
\colorlet{c26}{Tan!30}
\colorlet{c27}{TealBlue!30}

\newcommand{\NUM}{\texttt{[NUM]}}

\usepackage[accepted]{icml2025}

\usepackage{amsmath}
\usepackage{amssymb}
\usepackage{amsthm}

\usepackage[capitalize,noabbrev]{cleveref}

\theoremstyle{plain}

\theoremstyle{definition}

\theoremstyle{remark}

\usepackage[textsize=tiny]{todonotes}

\icmltitlerunning{From Tokens to Numbers: Continuous Number Modeling for SVG Generation}

\begin{document}

\twocolumn[
\icmltitle{From Tokens to Numbers: Continuous Number Modeling for SVG Generation}




\icmlsetsymbol{intern}{*}

\begin{icmlauthorlist}
\icmlauthor{Michael Ogezi}{uw,vector,canva,intern}
\icmlauthor{Martin Bell}{canva}
\icmlauthor{Freda Shi}{uw,vector}
\icmlauthor{Ethan Smith}{canva}
\end{icmlauthorlist}

\icmlaffiliation{uw}{Cheriton School of Computer Science, University of Waterloo, Waterloo, ON, Canada}
\icmlaffiliation{canva}{Canva, Sydney, NSW, Australia}
\icmlaffiliation{vector}{Vector Institute}

\icmlcorrespondingauthor{Michael Ogezi}{\href{mailto:mogezi@uwaterloo.ca}{mogezi@uwaterloo.ca}}

\icmlkeywords{Machine Learning, ICML, Number Representations, SVG Generation, Reinforcement Learning}

\vskip 0.3in
]



\printAffiliationsAndNotice{{\textsuperscript{*}This project was carried out during an internship at Canva with Ethan and Martin.}}  

\begin{abstract}
For certain image generation tasks, vector graphics such as Scalable Vector Graphics (SVGs) offer clear benefits such as increased flexibility, size efficiency, and editing ease, but remain less explored than raster-based approaches. 
A core challenge is that the numerical, geometric parameters, which make up a large proportion of SVGs, are inefficiently encoded as long sequences of tokens. 
This slows training, reduces accuracy, and hurts generalization. 
To address these problems, we propose {Continuous Number Modeling} {(CNM)}, an approach that directly models numbers as first-class, continuous values rather than discrete tokens. 
This formulation restores the mathematical elegance of the representation by aligning the model's inputs with the data's continuous nature, removing discretization artifacts introduced by token-based encoding.
We then train a multimodal transformer on 2 million raster-to-SVG samples, followed by fine-tuning via reinforcement learning using perceptual feedback to further improve visual quality. 
Our approach improves training speed by over 30\% while maintaining higher perceptual fidelity compared to alternative approaches. 
This work establishes CNM as a practical and efficient approach for high-quality vector generation, with potential for broader applications. 
We make our code available at \href{http://github.com/mikeogezi/CNM}{\texttt{https://github.com/mikeogezi/CNM}}.
\end{abstract}


 
\section{Introduction}

Vector image formats such as Scalable Vector Graphics (SVGs) provide a compact, resolution-independent, and semantically structured representation of visual content.
They describe images as sequences of geometric primitives such as lines, curves, and shapes, which are parameterized by continuous coordinates, angles, and style attributes.
In contrast, raster formats such as PNG or JPEG encode fixed grids of pixel intensities, losing the structural and geometric abstractions inherent to vector formats.
Consequently, SVGs are preferred in domains that require precision, compositionality, and editability, including user-interface design, data visualization, and programmatic illustration.

Despite these advantages, most image generation research remains focused on raster formats.
A central challenge for vector generation is that serialized SVGs produce long token sequences dominated by floating-point literals, which encode geometric parameters.
When processed by transformer-based large language models (LLMs) or vision-language models (VLMs), these values are decomposed into multiple discrete tokens, leading to substantial inefficiency in sequence modeling.

\begin{table*}[h!]
\small
\centering
\caption{
Sequence length compression factors for different tokenization strategies on the SVG-Stack test set.
A practical example applying each strategy to a sample sequence is provided for reference.
}
\label{table:comparing_tokenization}
\renewcommand{\arraystretch}{1.5} 
\setlength{\tabcolsep}{8pt} 
\tableskip

\begin{tabular}{lZc|ll}
\toprule
 & & & \multicolumn{2}{c}{\emph{Example} \quad $\cdots$\texttt{M 123.456 234.567}$\cdots$} \\
\textbf{Tokenizer} & {Dataset Tokens} & \makecell[c]{Compression\\Ratio} & {Tokens} & {Count} \\
\midrule

Default & 7.46M & $1.00\times$ &
\colorbox{c1}{\texttt{M}}\hspace{1pt}\colorbox{c2}{\texttt{1}}\hspace{1pt}\colorbox{c3}{\texttt{2}}\hspace{1pt}\colorbox{c4}{\texttt{3}}\hspace{1pt}\colorbox{c5}{\texttt{.}}\hspace{1pt}\colorbox{c6}{\texttt{4}}\hspace{1pt}\colorbox{c7}{\texttt{5}}\hspace{1pt}\colorbox{c8}{\texttt{6}}\hspace{1pt}\colorbox{c9}{\texttt{\ }}\hspace{1pt}\colorbox{c3}{\texttt{2}}\hspace{1pt}\colorbox{c4}{\texttt{3}}\hspace{1pt}\colorbox{c6}{\texttt{4}}\hspace{1pt}\colorbox{c5}{\texttt{.}}\hspace{1pt}\colorbox{c7}{\texttt{5}}\hspace{1pt}\colorbox{c8}{\texttt{6}}\hspace{1pt}\colorbox{c10}{\texttt{7}} 
& 16 \\

Number-aware & 4.62M & $1.61\times$ &
\colorbox{c1}{\texttt{M}}\hspace{1pt}\colorbox{c11}{\texttt{123}}\hspace{1pt}\colorbox{c5}{\texttt{.}}\hspace{1pt}\colorbox{c12}{\texttt{456}}\hspace{1pt}\colorbox{c9}{\texttt{\ }}\hspace{1pt}\colorbox{c13}{\texttt{234}}\hspace{1pt}\colorbox{c5}{\texttt{.}}\hspace{1pt}\colorbox{c14}{\texttt{567}} 
& 8 \\

Ours & 1.60M & 4.67$\times$ &
\colorbox{c1}{\texttt{M}}\hspace{1pt}\colorbox{c15}{\texttt{123.456}}\hspace{1pt}\colorbox{c16}{\texttt{234.567}} 
& 3 \\

\bottomrule
\end{tabular}
\end{table*}

For instance, the literal \texttt{123.456} is decomposed by modern open-weight tokenizers such as Qwen's \cite{wang2024qwen2vlenhancingvisionlanguagemodels, qwen2025qwen25technicalreport, yang2025qwen3technicalreport} into seven tokens, one per digit plus the decimal point (Table~\ref{table:comparing_tokenization}).
This fragmentation inflates sequence length, memory usage, and training cost.
Moreover, treating continuous coordinates as discrete text tokens is an artificial abstraction.
While suitable for mathematical reasoning tasks, such digit-level tokenization is poorly aligned with the structural regularities of SVG data, where nearby numeric values are highly correlated.
The resulting redundancy hinders generalization and reduces output fidelity.

We address these issues with a two-stage framework. 
First, we introduce {Continuous Number Modeling (CNM)}, in which each SVG is decomposed into two sequences: (i) a \emph{token sequence} that preserves symbolic tokens while replacing numeric literals with \NUM{} placeholder tokens, and (ii) a \emph{float sequence} that stores the corresponding original scalar values.
A dedicated \component{Number Encoder}, implemented with a Fourier feature map followed by a multi-layer perceptron (MLP), projects the numbers into the backbone transformer's input embedding space.
On the other end of the model, a \component{Number Decoder} reconstructs these continuous values from the corresponding last hidden state.
This approach substantially shortens input sequences while preserving numerical precision. 
Second, after supervised training, we apply reinforcement learning (RL) from perceptual feedback using Group Relative Policy Optimization \citep[GRPO;][]{shao2024deepseekmathpushinglimitsmathematicalgrpo}. We formulate a composite perceptual reward that combines SSIM \cite{1284395ssim}, LPIPS \cite{zhang2018unreasonableeffectivenessdeepfeatureslpips}, and DINOv2 similarity \cite{oquab2024dinov2learningrobustvisual}.
This stage further improves visual fidelity and alignment with human perceptual metrics.

On the SVG-Stack \cite{Rodriguez_2025_CVPR} benchmark, our approach models considerably shorter sequences, training performance by {32\%}, while also yielding consistent gains over the strongest baseline on multiple metrics.
Our contributions are summarized as follows:

\begin{enumerate}[leftmargin=*,topsep=0pt,itemsep=0pt]
    \item \textbf{Continuous number encoding for SVGs.}
    We propose a novel number representation format which directly encodes floating points as first-class, continuous values embedded via a Fourier mapping, avoiding the fragmentation and precision loss of discrete tokenization.


    \item \textbf{Comparison of various numerical representations.}
    We compare our continuous modeling approach to tokenization and quantization baselines on a fixed data budget, observing efficiency and performance gains.

   \item \textbf{The SVGFloat file format.}
    We propose a binary alternative to the SVG file format that persists geometric parameters as raw numerics, enabling compact storage.
\end{enumerate}

The remainder of this paper is organized as follows.
Section~\ref{sec:method} presents the proposed CNM method.
Section~\ref{sec:exp} details experimental design, datasets, evaluation metrics, and ablations.
Section~\ref{sec:related} reviews related work.
Section~\ref{sec:conclusion} concludes.


\section{Continuous Number Modeling (CNM)}
\label{sec:method}

\begin{figure*}[ht]
  \centering
  \includegraphics[width=\linewidth]{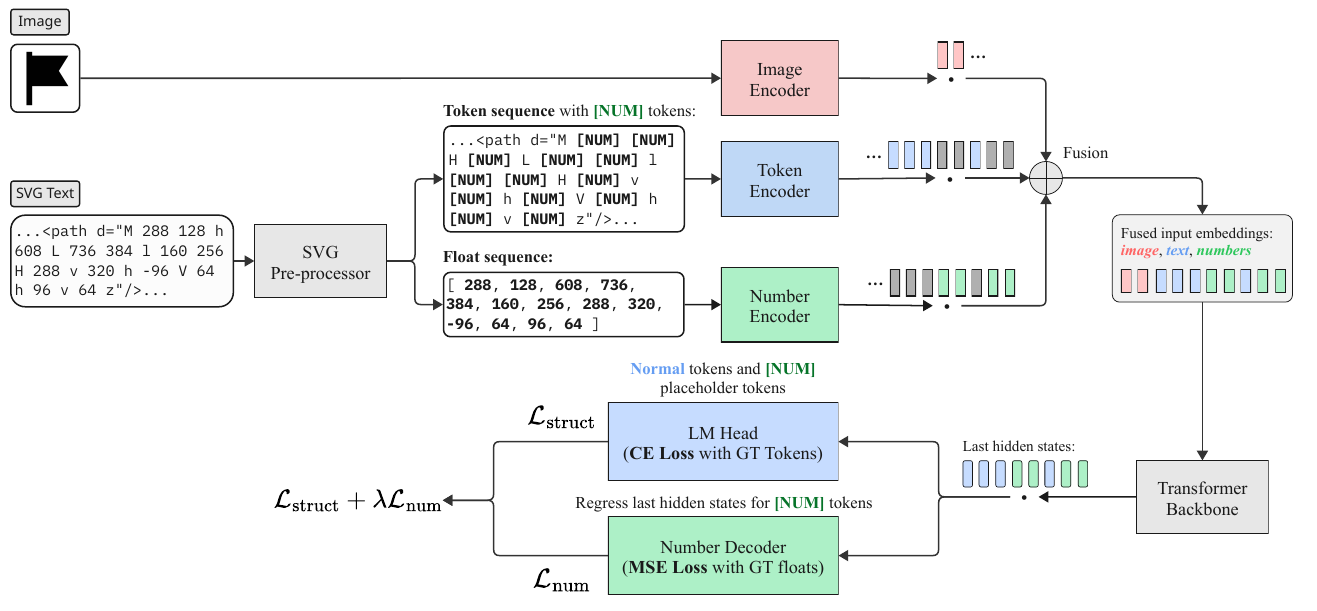}
  \caption{
      The training architecture for CNM:
      We train on raster-SVG pairs.
      The SVG path is split into two sequences: a token sequence with \NUM placeholders and a float sequence.
      Both sequences are embedded and fused, but maintain two distinct losses mediated by $\lambda$: cross-entropy for the tokens and MSE for the floats.
      Detailed expansions of the \component{Number Encoder} (Figure~\ref{fig:number_encoder} in Appendices) and \component{Number Decoder} (Figure~\ref{fig:number_decoder} in Appendices) are provided.
  }
  \label{fig:training}
\end{figure*}

\begin{figure*}[ht]
  \centering
  \includegraphics[width=0.80\linewidth]{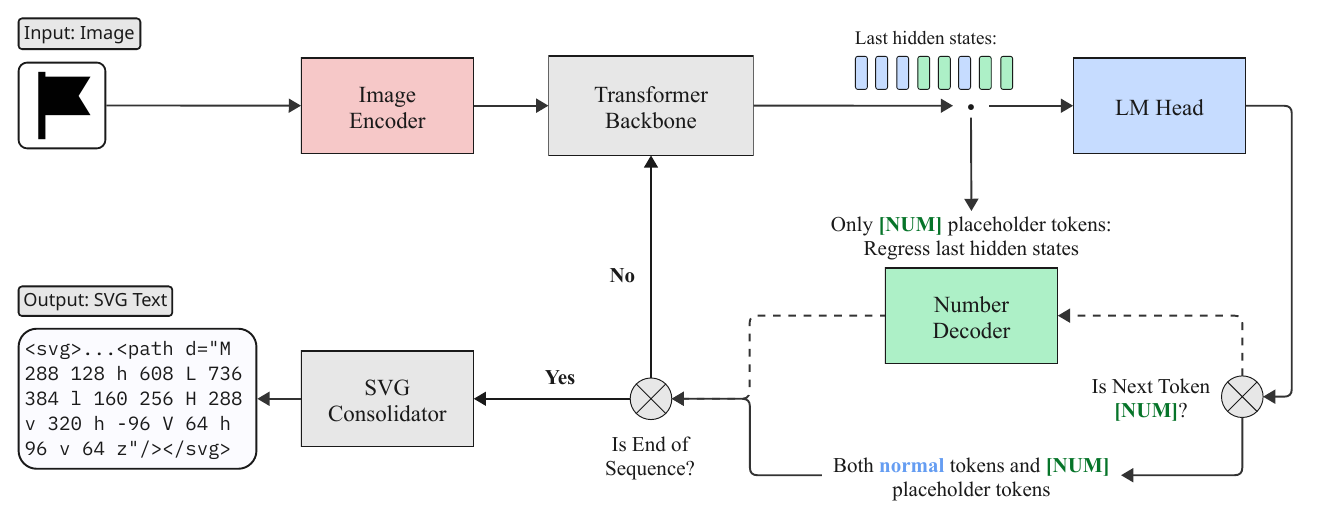}
  \caption{
      The inference architecture for CNM:
      A raster image is provided as input, and we generate \NUM and normal tokens.
      When a \NUM is encountered, we decode its last hidden state to a scalar value and add it to a float buffer.
      This value is re-embedded for the next step.
      At the end of the generation, we consolidate the normal tokens and the float values to produce a final SVG.
  }
  \label{fig:inference}
\end{figure*}

We propose a method for raster-to-vector generation where we treat numerical values within SVGs as first-class continuous variables. 
While prior work generically tokenizes numbers \cite{Rodriguez_2025_CVPR} or quantizes them into a limited number of discrete bins \cite{yang2025omnisvg}, we isolate and model each numeric literal directly. 
This approach enables us to disentangle an SVG's symbolic structure from its continuous parameters, allowing our model to learn the distinct properties of these elements better.
To implement our method, we augment a multimodal large language model (MLLM) with four key components: an \component{SVG Pre-processor}, a \component{Number Encoder} (composed of a \subcomponent{Normalizer}, a \subcomponent{Fourier Feature Mapping}, and an \subcomponent{MLP}), a \component{Number Decoder} (composed of an \subcomponent{MLP} and a \subcomponent{De-normalizer}), and an \component{SVG Consolidator}. 
These components can be seen in Figures~\ref{fig:training} and~\ref{fig:inference}.
We then train the model in two stages: first, we apply a supervised fine-tuning (SFT) objective, and second, we fine-tune it with an RL
objective to maximize the perceptual quality of the rendered output. 
The following subsections detail our architecture and training procedure.

\subsection{Overview}
\label{subsec:overview}

We frame vector generation as two parallel sub-tasks: (1) \emph{structural prediction} and (2) \emph{numerical prediction}.
On the one hand, the structural prediction task focuses on learning the syntax of SVGs. 
This entails the sequence of commands that compose an SVG, independent of numerical attributes such as specific coordinates, dimensions, or scales.
On the other hand, the numerical prediction task focuses exclusively on regressing these numerical attributes for a given structure. 
This disentanglement allows us to model each component with a specialized objective.


\paragraph{Structural prediction.} 
Here, the model autoregressively generates the sequence of SVG tokens. 
Its vocabulary comprises all tokens from the original model's vocabulary, denoted by $\mathcal{V}$, plus our special numerical placeholder token: \NUM.
Thus, given a target token sequence $y_{1:T}$ and an optional conditioning input $\mathbf{x}$, the model predicts the sequence as follows:
\[
\hat{y}_{1:T} = \arg\max_{y_{1:T}} \prod_{t=1}^{T} p(y_t \mid y_{<t}, \mathbf{x}).
\]
The loss for structural prediction is standard cross-entropy (CE):
\[
\mathcal{L}_{\text{struct}} = -\sum_{t=1}^{T} \log p(y_t \mid y_{<t}, \mathbf{x}).
\]
\paragraph{Numerical prediction.}
We use a simple regression to predict numbers.
At each timestep, $i$ where the model generates a \NUM token, we take the transformer's corresponding last hidden state, $h_{\text{last}, i} \in \mathbb{R}^d$, then feed it into a dedicated \component{Number Decoder}, $f_\theta$, that subsequently outputs a normalized scalar value, $\hat{v}_i$.
Next, we de-normalize $\hat{v}_i$ using the canvas bound $M$ to produce the final numerical attribute $\hat{n}_i$:
\[
\hat{v}_i = f_\theta(h_{\text{last}, i}), \quad \hat{n}_i = M \cdot \hat{v}_i.
\]
To train numerical prediction, we use a mean-squared error (MSE) loss
between the predicted and ground-truth values. 
The loss is averaged over all positions in the sequence where a number is predicted, denoted by the set $\mathcal{I} = \{ t_i : y_{t_i} = \NUM \}$:
\[
\mathcal{L}_{\text{num}} = \frac{1}{|\mathcal{I}|} \sum_{i \in \mathcal{I}} \left( \hat{n}_i - n_i \right)^2.
\]
\paragraph{Joint objective.}
Finally, we combine the structural and numerical losses into a joint objective function. 
We train the model end-to-end by minimizing this weighted sum:
\[
\mathcal{L} = \mathcal{L}_{\text{struct}} + \lambda \mathcal{L}_{\text{num}},
\]
where $\lambda$ is a hyperparameter that balances the contribution of the two task losses.
Empirically, we find that $\lambda$ must remain small (in the range of $10^{-5}$ to $10^{-3}$) for stable training since the MSE tends to be much larger than the CE.

With our dual-task framework established, the following subsections describe the components that bring it to life, beginning with our number pre-processing pipeline.

\subsection{Number Pre-Processing}
To prepare our data for training, each raw SVG string is converted into two parallel sequences: a sequence of tokens and a corresponding ordered sequence of continuous values.
To achieve this, we first parse the SVG path data, replace every floating-point literal with our special \NUM token, and store the original numeric values, in order, in a separate float sequence.
This process results in a token sequence that captures the SVG's pure syntax, alongside a float sequence with all its numerical attributes.
Finally, to prepare these numbers for the \component{Number Encoder} and \component{Number Decoder}, we normalize each value $n_i$ into a range of $[-1,1]$ by dividing it by the canvas-bound scalar $M$, such that $v_i=n_i/M$.
This entire transformation is applied to all data before both training and inference.
Our representation approach also motivates the \textsc{SVGFloat} file format, a binary storage format that mirrors this structure for efficient persistence; we experimentally demonstrate its compression efficiency in Section~\ref{par:svgfloat}.
\begin{figure*}[!h]
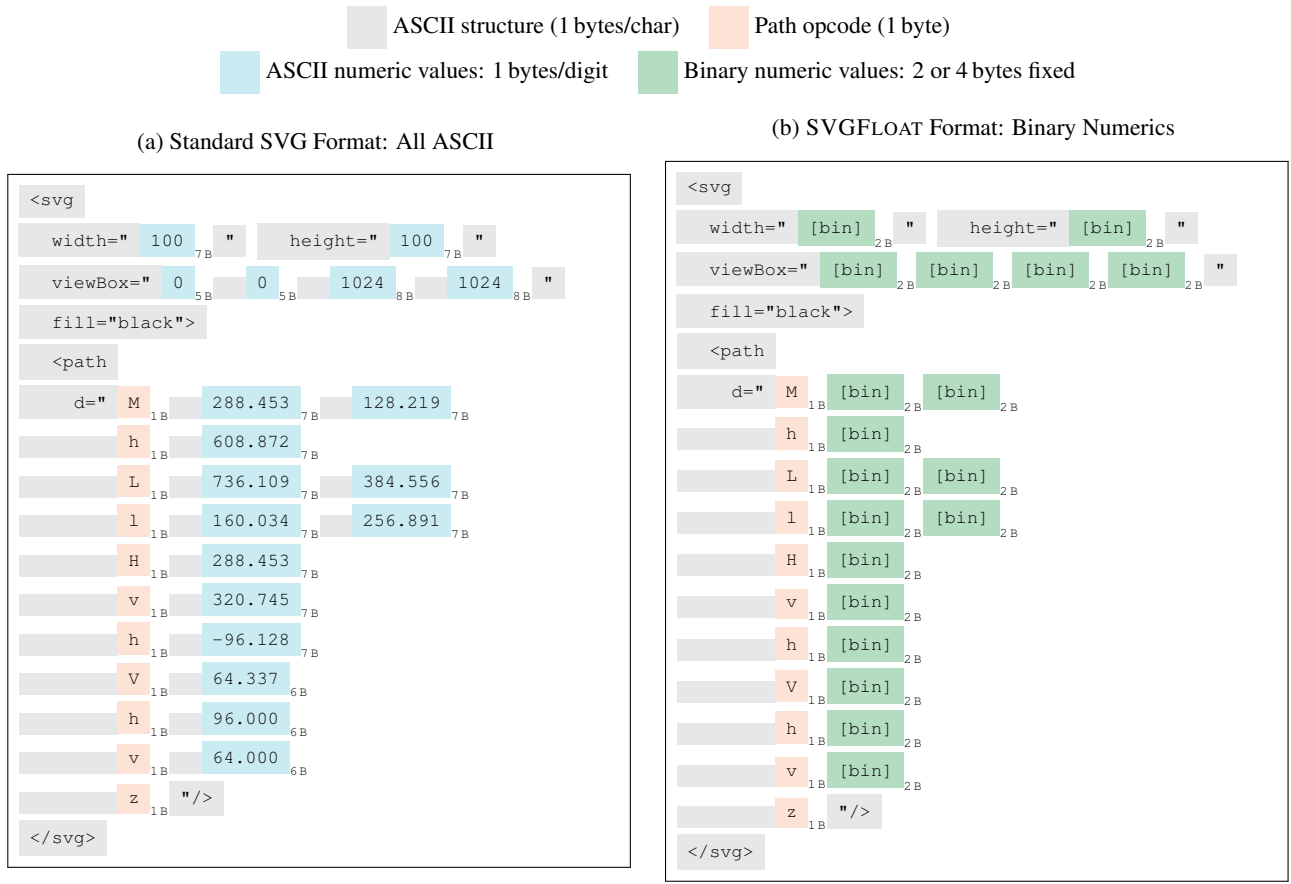

  \centering
  \small
  
  \caption{
      {Byte-level comparison of standard SVG vs.\ \textsc{SVGFloat}.}
      In standard SVG, all content is ASCII (1 byte per character).
      In \textsc{SVGFloat}, structural syntax remains ASCII, but numeric values are stored as raw binary floats (4 bytes for \texttt{float32}, 2 bytes for \texttt{float16} or \texttt{bfloat16}).
      Bracketed segments \texttt{[bin]} indicate binary-encoded floats with fixed byte size regardless of decimal precision.
      Note: In \textsc{SVGFloat}, separators between consecutive numbers are omitted; the fixed byte width disambiguates boundaries.
  }
  \label{fig:svgfloat_format}
  
  \vspace{0.6em}
  
  \begin{center}
  {\small
  \colorbox{Gray!20}{\strut\phantom{xx}}~ASCII structure (1\,bytes/char) \quad
  \colorbox{c1!40}{\strut\phantom{xx}}~Path opcode (1\,byte) \quad
  \\
  \colorbox{SkyBlue!30}{\strut\phantom{xx}}~ASCII numeric values: 1\,bytes/digit
  \quad
  \colorbox{Green!30}{\strut\phantom{xx}}~Binary numeric values: 2 or 4\,bytes fixed
  }
  \end{center}
  
  \vspace{0.4em}
  
  \begin{minipage}{0.49\linewidth}
    \centering{(a) Standard SVG Format: All ASCII}
    \vspace{0.3em}
    \input{figs/svg}
  \end{minipage}
  \hfill
  \begin{minipage}{0.49\linewidth}
    \centering{(b) \textsc{SVGFloat} Format: Binary Numerics}
    \vspace{0.3em}
    \input{figs/svgfloat}
  \end{minipage}
  
  \end{figure*}

\paragraph{The SVGFloat Format.}
The \textsc{SVGFloat} format is a binary–text hybrid alternative to standard SVG storage that replaces textual numeric encoding with raw floating-point values while preserving the line-oriented structure of SVG.
In conventional SVG, coordinates are stored as variable-length ASCII strings (e.g., \texttt{``0.5''} vs.\ \texttt{``123.456789''}), which may be inefficient.
\textsc{SVGFloat} instead stores every numeric parameter as a {fixed-width binary float}, for a more compact representation.

Unlike the dual-sequence scheme we described in the main text, \textsc{SVGFloat} uses a single stream.
Each \textsc{SVGFloat} file is a single interlaced ASCII/binary stream.
Structural SVG syntax (element names, attributes, separators, and delimiters) is preserved in ASCII form, while all numeric values are written as raw binary floats in sequence order.

To preserve path continuity, we use {NaN-boxing}~\cite{Gudeman1993RepresentingTI} to encode both path commands and numerical values within a unified stream.
By exploiting the IEEE~754 NaN representation (where all exponent bits are set to 1, and the mantissa is non-zero), we store command opcodes in the mantissa bits of NaN values.
Each SVG path segment (e.g., \texttt{M}, \texttt{L}, \texttt{C}, \texttt{A}) is encoded as a NaN with its command type in the mantissa, followed by the numeric parameters written as regular floats.
This approach enables a fully homogeneous floating-point encoding where commands and coordinates coexist in a single stream without requiring separate opcode bytes.
Inspired by our work in machine learning, where mixed precision approaches are common, this format is not locked to a specific floating-point representation, supporting IEEE~754 \texttt{float32} as well as \texttt{float16} and \texttt{bfloat16}.

Compression efficiency and visual fidelity results are presented in Table~\ref{tab:svgfloat_results} in Section~\ref{sec:exp}. 
We will provide a reference reader and writer in our code release.

\subsection{Dataset Filtering and Pre-Processing}
\label{subset:filtering_and_preprocessing}

We use the SVG-Stack dataset \cite{Rodriguez_2025_CVPR}. 
We prepare this data with a multi-stage filtering and normalization pipeline designed to ensure two key properties: numerical stability, by bounding all coordinate values, and visual consistency, by ensuring our transformations do not alter the rendered appearance.
First, each SVG is simplified using the \texttt{SVGO} optimization tool.\footnote{\url{https://svgo.dev/}}
This initial step is crucial for applying all affine transformations directly to the path coordinates, which is essential for accurate scaling. 
Next, we implement a recursive scaling process to fit every SVG to a canonical canvas of size $[-M, M]^2$. This involves calculating a global scale factor and applying it to all numerical attributes, such as coordinates, lengths, and stroke widths. 
For robust handling of complex path commands, we use the \texttt{svg.path} library to parse,\footnote{\url{https://pypi.org/project/svg.path/}} scale, and reconstruct the path data.
Finally, to guarantee that our scaling process does not significantly alter the visual appearance of the graphic, we perform a quality control check. 
We render both the original and the transformed SVGs as raster images using the \texttt{cairosvg} library.\footnote{\url{https://cairosvg.org/}} 
We then calculate the structural similarity index (SSIM) between the two images. 
Any sample with an SSIM score below a strict threshold of $0.99$ is discarded, as is any sample containing a final numerical value that violates the canvas bound $|n_i| > M$. 
This rigorous process yields a clean, numerically stable, and visually aligned dataset that is fully compatible with our continuous modeling pipeline.

\subsection{Multimodal Embeddings}

Our model processes a heterogeneous input sequence containing symbolic tokens, continuous numerical values, and image features. 
We embed each element according to its modality, mapping all inputs into a shared $d$-dimensional latent space. 
This unified representation allows a single transformer decoder to process the entire sequence jointly.
We detail the approach for each modality below:

\paragraph{Token embeddings.}

We represent discrete tokens, such as SVG markup and path commands (e.g., \texttt{<svg}, \texttt{height=}, \emph{M}, \emph{L}, etc.), and delimiters, using a standard learned token embedding layer, $e_{\text{text}}$.
This also includes the placeholder token \NUM.
This layer maps each token from our vocabulary $\mathcal{V'}$ into a common input embedding space:
\[
e_{\text{text}}: \mathcal{V'} \rightarrow \mathbb{R}^d
\]

\paragraph{Number embeddings (with Fourier feature maps).}

Embedding a scalar value into a high-dimensional space can be non-trivial, and while a simple linear projection can be effective for embedding scalars \cite{golkar2024xvalcontinuousnumericaltokenization}, our task required a more expressive mapping. 
To create a more expressive representation, we use Fourier feature maps, a technique inspired by their success in positional encodings within transformers. 
This approach maps each normalized number $v \in [-1,1]$ into a higher-dimensional space using a set of sinusoidal functions with exponentially increasing frequencies. 
This makes the final representation adequately sensitive to variations in the input value, which is crucial for geometric precision and generalization.

First, we expand $v$ into a $2k$-dimensional feature vector, $\gamma(v)$:
\[
\begin{split}
    \gamma(v) = \big[ & \sin(2^0 \pi v), \cos(2^0 \pi v), \ldots, \\
                     & \sin(2^{k-1} \pi v), \cos(2^{k-1} \pi v) \big] \in \mathbb{R}^{2k}
\end{split}
\]
This feature vector is then projected into the backbone transformer's input embedding space by our \component{Number Encoder}, which is parametrized by an MLP and denoted as $g_\phi$.
This expansion layer maps the $2k$-dimensional input to the final $d$-dimensional number embedding:
\[
e_{\text{num}}(v) = g_\phi(\gamma(v)) \in \mathbb{R}^d
\]
\paragraph{Image embeddings.}
To condition the generative process on a raster image, $I$, we first encode it into the same $d$-dimensional latent space. 
For this, we use the ready-to-go vision encoder of our pretrained MLLM.
The vision encoder processes the input image by dividing it into a grid of non-overlapping patches and computing an embedding for each one. 
This results in a sequence of $L$ patch embeddings, which collectively represent the image's content:
\[
e_{\text{img}}(I) \in \mathbb{R}^{L \times d}
\]
\paragraph{Embedding fusion.}
Once the tokens, numbers, and image patches are mapped to our shared $d$-dimensional latent space, we fuse them into a single sequence for the transformer decoder.
This fusion is achieved by concatenating the individual embedding vectors in their original sequence order. 
For example, the final input representation for the model might be constructed as follows:
\[
S_{\text{input}} = [ \underbrace{e_{\text{img}_1}, \ldots, e_{\text{img}_L}}_{\text{Image Patches}}, \underbrace{\ldots, e_{\text{text}_i}, e_{\text{num}_j}, \ldots}_{\text{SVG Tokens \& Numbers}} ]
\]
\subsection{Training Regularization}
To make the model robust to exposure bias, the discrepancy between seeing ground-truth values during training and its own predictions during inference, we regularize the training data. 
Specifically, for each normalized number value $v$, we create a noisy version, $\tilde{v}$ as follows:
\[
\begin{split}
    \tilde{v} = v + \epsilon, \quad \text{where} \quad \epsilon &\sim \mathcal{N}(0, \sigma^2_{\text{noise}}) \\
    \text{and} \quad \sigma_{\text{noise}} &= \frac{\eta_{\text{noise}}}{M}
\end{split}
\]
The hyperparameter, $\eta_{\text{noise}}$ controls the noise intensity. 
By defining the standard deviation $\sigma_{\text{noise}}$ relative to the canvas/normalization constant, $M$, the noise level is effectively specified in the more intuitive, absolute coordinate space.
This data augmentation enhances generalization by making the model more robust to the small prediction errors it will encounter during inference.

\subsection{Post-Processing and Reconstruction}

At inference, the model produces two outputs: a sequence of symbolic tokens containing \NUM{} placeholders, and an ordered sequence of corresponding normalized numbers. 
The \component{SVG Consolidator} merges these two streams to reconstruct a single, valid SVG string.

The process iterates through the token sequence. 
When a \NUM{} placeholder is encountered, the next available number, $\hat{v}_i$, is taken from the predicted sequence. 
This value is first de-normalized by the canvas scaling factor $M$:
\[
\hat{n}_i = M \cdot \hat{v}_i.
\]
Before being inserted into the string, the resulting scalar $\hat{n}_i$ is rounded to a fixed decimal precision and clipped to the canvas bounds $[-M, M]$.

\subsection{RL for Fidelity Improvement}
\label{subsec:reward_function}

To better align the model's output with perceptual quality, which supervised losses do not fully capture, we introduce an RL fine-tuning stage. 
We treat the generator as a policy $\pi_\theta$ and optimize it to maximize a perceptual reward $R$. This reward is a weighted sum of SSIM, LPIPS', and DINOv2 similarity:
\[
R(I_{gt}, I_{pred}) = \alpha \cdot \mathrm{DINOv2_{sim}} + \beta \cdot \mathrm{SSIM} + \gamma \cdot \mathrm{LPIPS'}
\]
We use multiple perceptual metrics to boost robustness.
Also, note that LPIPS' is a similarity rather than a distance.
We also force all components to be $[0,1]$. 
We use Group Relative Policy Optimization (GRPO), a policy gradient method that estimates the advantage $\hat{A}_t$ by comparing multiple sampled outputs from the policy. 
The policy is then updated using this advantage-weighted gradient:
\[
\nabla_\theta J(\theta) \approx \mathbb{E}_{\pi_\theta} \left[ \hat{A}_t \nabla_\theta \log \pi_\theta(a_t \mid s_t) \right]
\]
In summary, our approach first uses Continuous Number Modeling to improve geometric precision, then applies RL to directly optimize for perceptual quality. 
The following experiments validate its effectiveness.

\section{Experiments}
\label{sec:exp}

To evaluate the effectiveness of Continuous Number Modeling and RL fine-tuning, we conduct a comprehensive set of experiments. 
We compare our method against state-of-the-art baselines on the SVG-Stack benchmark and perform detailed ablation studies to analyze the contribution of each component of our approach.

\subsection{Experimental Setup}

\begin{table*}[!t]
 \setlength{\tabcolsep}{2.5pt}
 \small
 \centering
 \caption{
    Main results on the SVG-Stack test set. All methods use the same Qwen2-VL-2B backbone for a fair comparison.
    Higher is better ($\uparrow$) for SSIM, DINOv2, LPIPS', and CC/SL; lower is better ($\downarrow$) $ \sqrt{\text{MSE}} $, Runtimes, Seq. Len. (Sequence Length), and Char. (Character) Count. 
    Best results are in \textbf{bold} except for Non-ML approaches, which are present for reference purposes.
    LPIPS' is a similarity score rather than a distance, so higher is better.
    LPIPS', SSIM, and DINOv2 are expressed as percentages.
    Train/Test Runtime are based on time per step during training and (normalized) total generation time, respectively.
    CC/SL is a compression ratio from Char. Count divided by Seq. Length.
 }
 \label{tab:main_results}
 \tableskip
 
 \begin{tabular}{llccccZccRRR}
  \toprule
  & \textbf{Method} 
& \makecell{SSIM \\ ($\uparrow$)}
& \makecell{LPIPS' \\ ($\uparrow$)}
& \makecell{DINOv2 \\ ($\uparrow$)}
& \makecell{$\sqrt{\text{MSE}}$ \\ ($\downarrow$)}
& \makecell{Human \\ Prefs. \\ ($\uparrow$)}
& \makecell{\textit{Train} \\ Runtime (s) \\ ($\downarrow$)}
& \makecell{\textit{Test} \\ Runtime (s) \\ ($\downarrow$)}
& \makecell{\textit{Test} \\ Seq. Len. \\ ($\downarrow$)}
& \makecell{\textit{Test} \\ Char. Count \\ ($\downarrow$)}
& \makecell{\textit{Test} \\ CC/SL \\ ($\uparrow$)} \\
  \cmidrule(lr){2-12} 

  \multirow{2}{*}{\emph{Ours}}
  & {SFT-only}   & 54.1 & 45.3 & 42.5 & 82.4 & -- & {0.71} & 0.47 & 536 & 2,893 & {5.40} \\ 
  & SFT+RL & 55.9 & 46.0 & 44.8 & 79.6 & -- & 3.07 & 0.44 & 549 & 2,478 & 4.51 \\
  \midrule
  
  \multirow{2}{*}{\emph{Training-based}} 
  & StarVector          & 48.5 & 44.5 & 42.0 & 58.4 & -- & 1.04 & 0.51 & 1,189 & 1,471 & 1.24 \\ 
  & OmniSVG             & 46.0 & 40.3 & 32.6 & 87.3 & -- & 0.79 & 0.58 & 670 & 994 & 1.48 \\ 
  \midrule

  \multirow{1}{*}{\emph{Zero-shot}}
  & Qwen2VL-2B (base) & 13.6 & 12.0 & 7.0 & 61.7 & -- & -- & 0.50 & {46} & {103} & 2.24 \\
  \midrule
  
  \multirow{2}{*}{\emph{Non-ML}}
  & Potrace             & 84.5 & 81.8 & 73.8 & 61.7 & -- & -- & 0.02 & -- & 3,027 & -- \\
  & VTracer             & 95.3 & 91.2 & 90.2 & 39.1 & -- & -- & 0.26 & -- & 38,304 & -- \\
  \bottomrule

 \end{tabular}
\end{table*}

\paragraph{Datasets.}

We train our model and relevant baselines on the train split of the SVG-Stack dataset~\cite{Rodriguez_2025_CVPR} and evaluate on the test split. 
We train with 3 different seeds and report the mean.
After applying the filtering process described in Section \ref{subset:filtering_and_preprocessing}, the dataset comprises 2.17M training samples, 108K validation samples, and 5.71K test samples. 


\paragraph{Baselines.}
We compare against two training-based baselines: StarVector~\cite{Rodriguez_2025_CVPR}, which uses vanilla tokenization, and OmniSVG~\cite{yang2025omnisvg}, which uses quantization.
Same as ours, both use the \texttt{Qwen2-VL-2B-Instruct} backbone and are trained on the same dataset for an equal number of epochs, placing all trained models in a matched parameter and data regime.
We note that the primary distinction between our approach and these others lies in how numerical values are represented and predicted. 
We additionally compare against classical, non-ML raster-to-vector systems (VTracer\footnote{\url{https://github.com/visioncortex/vtracer}}, Potrace\footnote{\url{https://potrace.sourceforge.net}})
and the base \texttt{Qwen2-VL-2B-Instruct}~\cite{wang2024qwen2vlenhancingvisionlanguagemodels} model.
See the full comparison in Table~\ref{tab:main_results}.

\paragraph{Evaluation metrics and details.}

We evaluate all methods by measuring the perceptual similarity between the ground-truth raster images and the rendered outputs of the generated SVGs. 
Specifically, we use SSIM (Structural Similarity Index)~\cite{1284395ssim} for structural fidelity and DINOv2 cosine similarity~\cite{oquab2024dinov2learningrobustvisual} for high-level semantic correspondence.
We also use LPIPS (Learned Perceptual Image Patch Similarity)~\cite{zhang2018unreasonableeffectivenessdeepfeatureslpips} with an AlexNet backbone, but in our evaluations, we report LPIPS', which denotes similarity rather than distance, resulting in higher being better. 
Additionally, we report $ \sqrt{\text{MSE}} $ for coordinate-level regression error, Seq. Length or Char. Count to quantify sequence length, and (normalized) Train/Test Runtimes to measure training/inference latencies.
For the main results in Table~\ref{tab:main_results}, we report strict variants of our metrics.
Under the lenient setting, we sanitize the generated SVG and, as a last resort, fall back to a default SVG if the output is invalid.
Under the strict setting, any invalid SVG is assigned a score of 0.

During inference, we sample with $temperature = 0.6$, $top\_p = 0.95$, and $top\_k = 5$, up to $2{,}048$ tokens.

\subsection{Implementation Details}

\paragraph{Model architecture.}

We implement our model by augmenting the \texttt{Qwen2-VL-2B-Instruct} architecture. 
Our custom modules are the \component{Number Encoder} $g_\phi$ and \component{Number Decoder} $f_\theta$.
The \component{Number Encoder} consists of a \subcomponent{Normalizer} (to which we add noise during training) followed by a \subcomponent{Fourier Feature Mapping} (with $k=16$ bands), and a multi-layer perceptron (MLP) (with GELU activations and $2$ hidden layers).
The \component{Number Decoder} consists of an MLP (with the same architecture as the \component{Number Encoder}'s MLP) followed by a \subcomponent{De-normalizer}.
As noted, all SVGs are normalized to a canvas of $[-512, 512]^2$, setting our normalization constant $M=512$.

\paragraph{SFT.}

We first train the model using our joint supervised objective. 
We use the AdamW optimizer with a cosine learning rate schedule and a base learning rate of $2 \times 10^{-5}$. 
We use an effective batch size of $128$ and train for $2$ epochs. 
To optimize training efficiency, the transformer backbone is trained in \texttt{bfloat16} precision, while our \component{Number Encoder} and \component{Number Decoder} operate in \texttt{float32}. 
Other key hyperparameters for this stage include the numerical loss weight $\lambda=1 \times 10^{-5}$ and the standard deviation of the injected Gaussian noise, $\sigma_{\text{noise}}=0.2$.
We find that training the entire model in \texttt{float32} yields no discernible performance benefit while \emph{significantly increasing} training time, and thus stick to the aforementioned \texttt{bfloat16} + \texttt{float32} format.

\paragraph{Reinforcement learning with perceptual feedback.}

The second stage applies RL via GRPO \cite{shao2024deepseekmathpushinglimitsmathematicalgrpo} to fine-tune the supervised-fine-tuned policy. 
For each input image, we sample $8$ candidate SVGs with $temp=1$, $top\_k=50$, and $top\_p=0.95$.
The perceptual reward combines three terms with weights $\alpha=0.4$, $\beta=0.3$, and
$\gamma=0.3$.
We use an effective batch size of $2048$ over $200$ optimization steps with a learning rate of $1\times10^{-6}$. 
The update is constrained by a target KL-divergence of $0.001$ relative to the reference model.

\paragraph{SVGFloat file format.}
\label{par:svgfloat}

Beyond its benefits for model training and inference, our representation also enables a more efficient storage format for SVG files.
Table~\ref{tab:svgfloat_results} shows the compression efficiency of \textsc{SVGFloat} (see Section~\ref{sec:method} for implementation details).
Our \texttt{float16} variant achieves a \textbf{42\%} reduction in file size compared to raw SVGs and \textbf{19\%} compared to gzipped SVGs, with negligible loss in visual fidelity.
This validates that our approach extends beyond model architecture to provide practical benefits for SVG storage and transmission.

\begin{table}[!t]
 \setlength{\tabcolsep}{6pt}
 \centering
 \caption{
    Comparison of compression efficiency and visual fidelity for \textsc{SVGFloat} variants against a gzipped SVG baseline.
    The \texttt{float16} format provides an optimal trade-off between efficient compression and high fidelity.
 }
 \label{tab:svgfloat_results}
 \tableskip
 \footnotesize
 \begin{tabular}{ llccc }
  \toprule
  & & \multicolumn{2}{c}{{Compression Ratio} $\uparrow$} & \\
  \cmidrule(lr){3-4}
  & \textbf{Format} & {vs.\ Raw} & {vs.\ GZIP} & {SSIM} $\uparrow$ \\
  
  \midrule

  \multirow{1}{*}{\centering \emph{Baseline}} 
  & ASCII SVG & $1.00\times$ & $1.00\times$ & 1.000 \\
  
  \midrule

  \multirow{3}{*}{\centering \emph{SVGFloat}}
  & \texttt{float32} & $1.14\times$ & $0.91\times$ & 0.986 \\
  & \texttt{float16} & {1.73$\times$} & {1.23$\times$} & {0.985} \\
  & \texttt{bfloat16} & $1.73\times$ & $1.31\times$ & 0.962 \\

  \bottomrule
 \end{tabular}
\end{table}




\subsection{Main Results}
We present our main quantitative results on the SVG-Stack test set in Table~\ref{tab:main_results}.
The results validate the effectiveness of CNM, demonstrating that treating geometric parameters as continuous values yields a concise and structurally efficient representation without sacrificing perceptual fidelity.

\paragraph{Efficiency and sequence compression.}
The most significant benefit of our approach is the considerable reduction in sequence length.
Compared to the strongest baseline, StarVector, our model reduces the average test sequence length from 1,189 to 549 tokens, a reduction of \textbf{54\%}.
As a result, during the SFT stage, our approach is considerably faster, reducing training runtime per batch from 1.04s (StarVector) to 0.71s, a \textbf{32\%} speedup.
Further, during inference, we also spend less time (0.47s) vs. StarVector's 0.51s, a \textbf{8\%} speedup, even with the overhead of custom modules, since we use fewer steps.

\paragraph{Perceptual and semantic fidelity.}
Despite using less than half the tokens of the baseline, our method achieves state-of-the-art results on key fidelity metrics.
Our approach achieves the highest SSIM (\textbf{54.1\%}), surpassing StarVector (48.5\%) and significantly outperforming the quantization-based OmniSVG (46.0\%).
On high-level semantic correspondence measured by DINOv2, we achieve a score of \textbf{42.5\%}, beating StarVector (42.0\%) and others.
Further, Table~\ref{tab:validity_rates} shows that we generate first-time valid SVGs at higher rates than other approaches.

\begin{table}[h]
\centering
\small
\caption{Validity rates for generated SVGs}
\label{tab:validity_rates}
\begin{tabular}{lc}
\toprule
\textbf{Method} & {Validity Rate} (\%) \\
\midrule
SFT (Ours) & 66.2 \\
StarVector & 55.0 \\
OmniSVG & 60.9 \\
\bottomrule
\end{tabular}
\end{table}

\paragraph{The impact of RL.}
The comparison between SFT-only and SFT+RL highlights the critical role of the RL fine-tuning stage.
While SFT alone yields a highly efficient model, the RL stage bridges the gap in perceptual alignment.
The RL update significantly boosts DINOv2 similarity (from 42.5\% $\rightarrow$ 44.8\%) and LPIPS' (from 45.3\% $\rightarrow$ 46.0\%), validating the use of a composite reward function to refine visual fidelity beyond what supervised training can achieve.

\paragraph{Baseline analysis.}
The limitations of alternative numerical representations are evident in the results.
OmniSVG's poor performance in SSIM (46.0\%) and DINOv2 (32.6\%) confirms that binning continuous coordinates into discrete codes severely limits geometric precision, especially without extensive training.
Meanwhile, StarVector achieves competitive visual scores but requires \textbf{2.1$\times$} more tokens to encode the same content, proving that standard LLM tokenizers are inefficient for dense geometric data.




\subsection{Ablation Studies}

To better understand and validate our key design choices, we conduct a series of ablation studies.
We show all our ablation results in Table~\ref{tab:ablations_merged}.
To start off, we analyze the core architectural choice: \textit{Number Representation}. 
Next, we look at the impact of changing the normalization constant $M$ during SFT. 
We then separately analyze the component \textit{sub-}rewards of our perceptual reward function for the RL stage (\textit{RL Reward Components}).
We also look at the effects of modifying the \textit{Encoding Mechanism}, \textit{Gaussian Noise}, and \textit{Precision Format}.

\begin{figure*}[!t]
 \centering
\caption{
    Qualitative comparison on the SVG-Stack test set.
    The figure contrasts our model (SFT + RL) with training-based baselines (StarVector, OmniSVG)
    .
    The ground-truth example contains thin strokes and small spatial offsets that require precise geometric prediction.
    Baselines either omit or distort these elements, whereas our method preserves stroke geometry and layout without over-smoothing.
    We provide more qualitative samples in Appendix \ref{app:qualitative}.
}
 \label{fig:qualitative_results}
 \tableskip
 \small
 \begin{tabular}{cccc
 }
  {\makecell{Ground\\Truth}} & {Ours} & {StarVector} & {OmniSVG} 
  \\
  
  \includegraphics[width=0.14\linewidth]{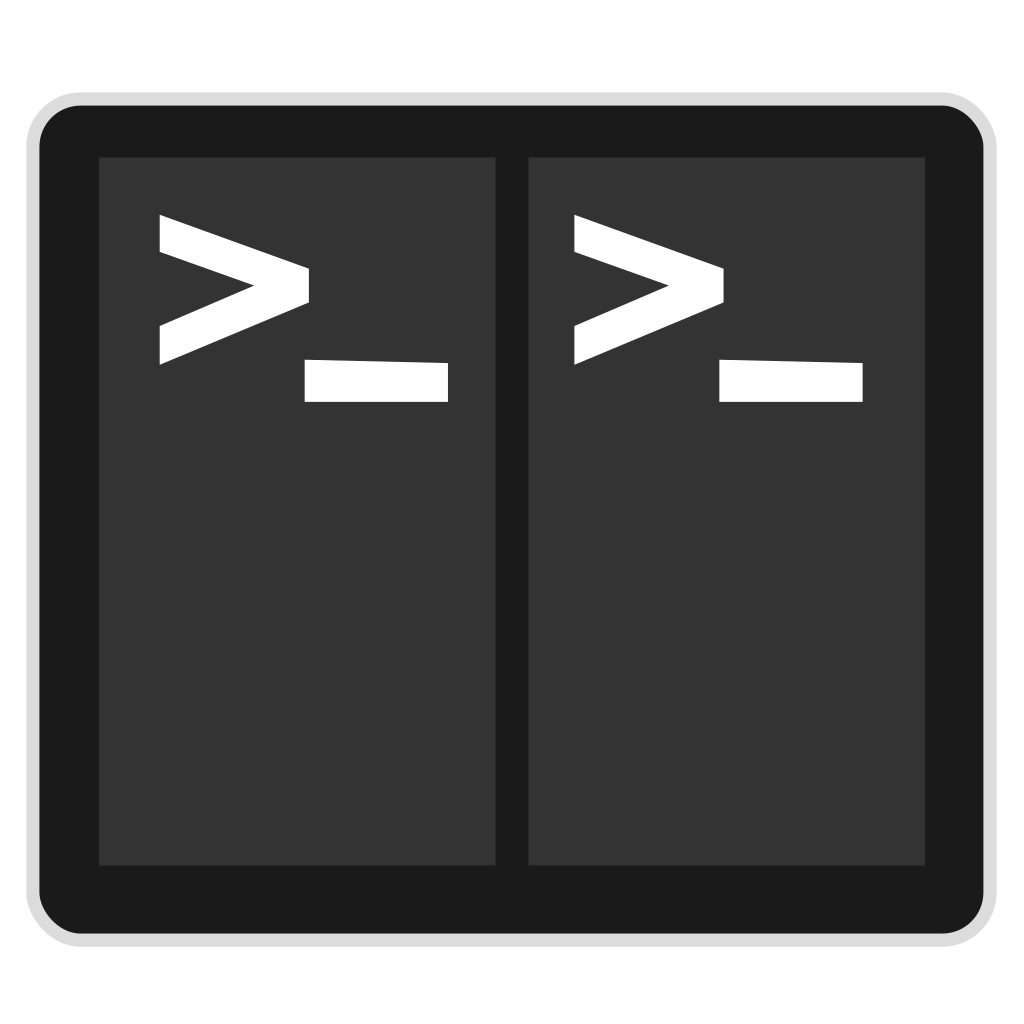} &
  \includegraphics[width=0.14\linewidth]{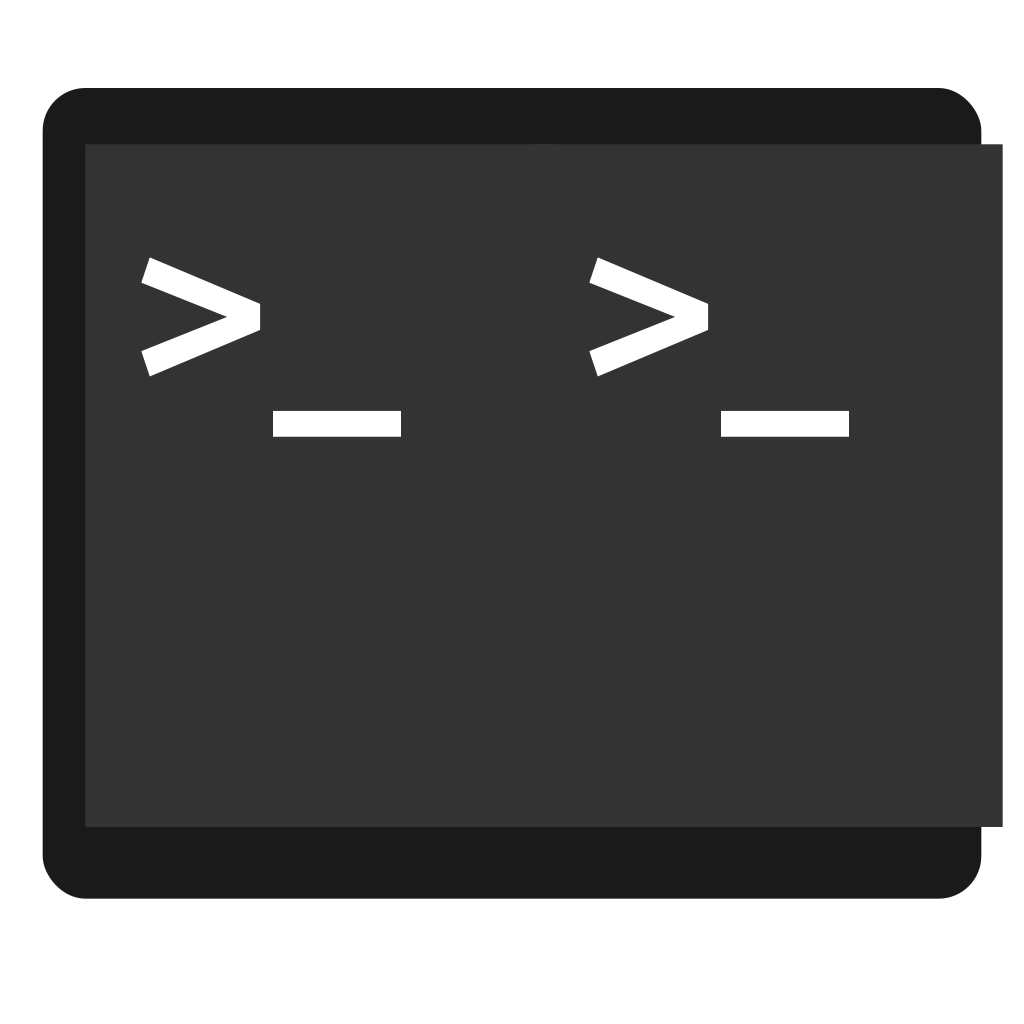} &
  \includegraphics[width=0.14\linewidth]{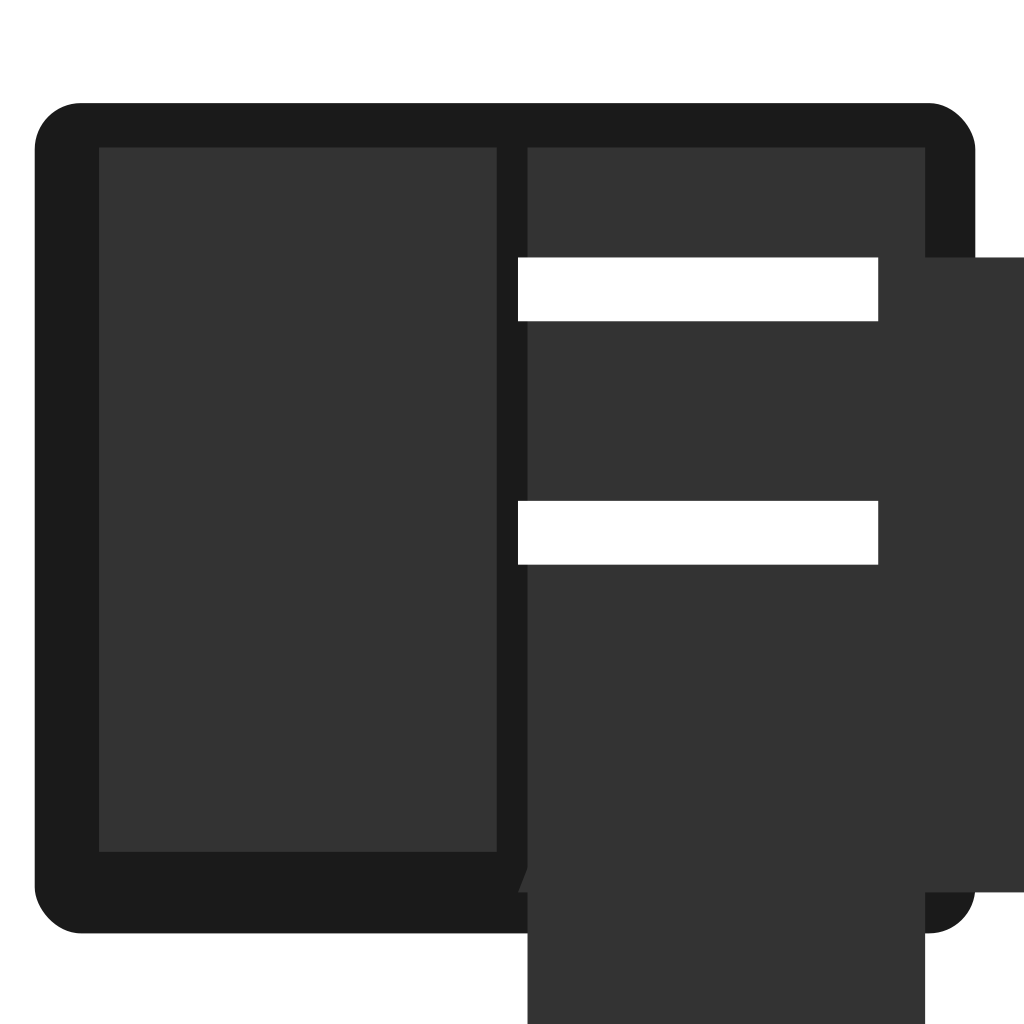} &
  \includegraphics[width=0.14\linewidth]{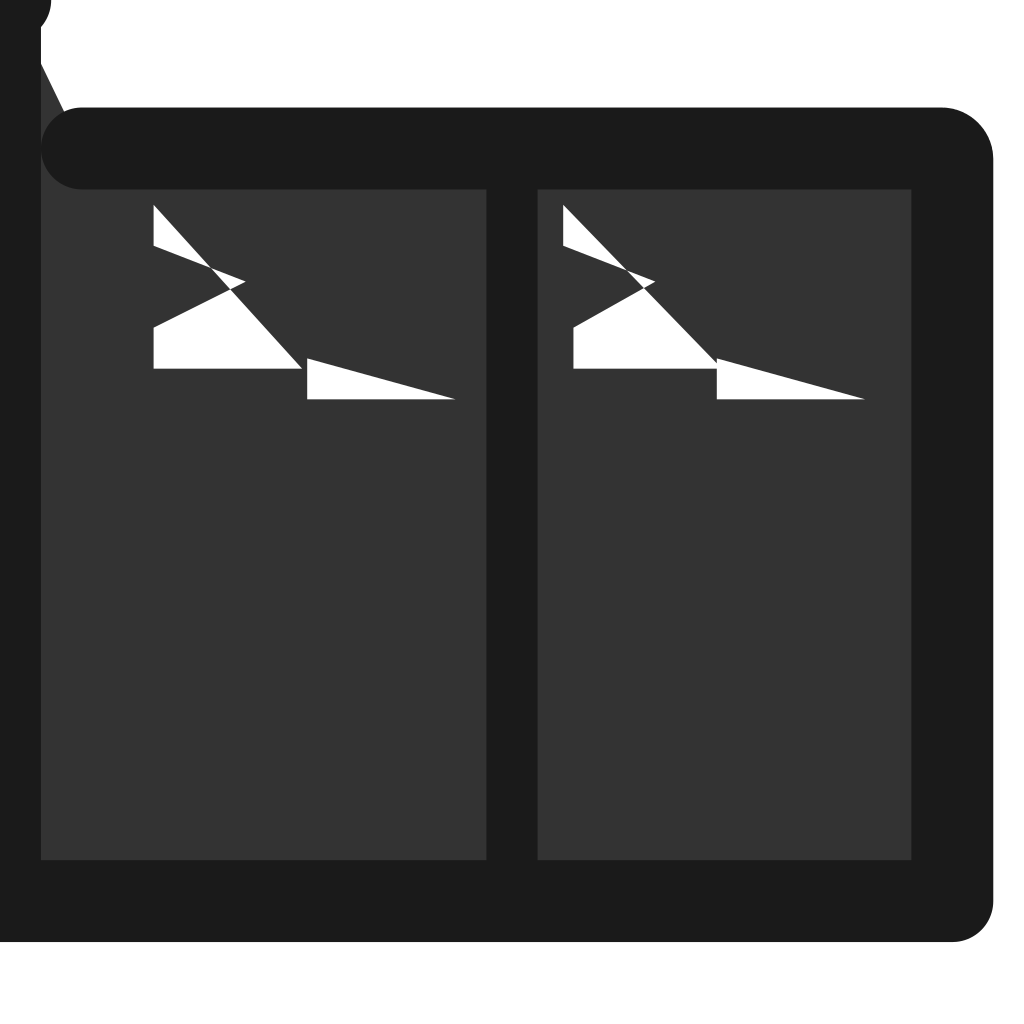} 
  \\
 \end{tabular}
\end{figure*}

\begin{table}[!t]
 \setlength{\tabcolsep}{3pt}
 \centering
 \caption{Ablation studies on core architectural choices, hyperparameters, and RL reward components on a subset of the SVG-Stack test set using the SSIM metric.}
 \label{tab:ablations_merged}
 \tableskip
 \footnotesize
 \begin{tabular}{ llcZ }
  \toprule
  & \textbf{Ablation} & \makecell{Lenient\\SSIM ($\uparrow$)} & {DINOv2 $\uparrow$} \\
  
  \midrule

  \multirow{3}{*}{\centering \emph{Number Representation}} 
  & Default Tokenizer & 0.53 & 52.1 \\
  & Quantized Numbers & 0.49 & 23.6 \\
  & Ours (Continuous) & 0.53 & 57.8 \\
  
  \midrule

  \multirow{2}{*}{\centering \emph{Encoding Mechanism}}
  & Fourier Feature Mapping & 0.53 & -- \\
  & Linear Projection & 0.44 & -- \\
  
  \midrule
  
  \multirow{3}{*}{\centering \emph{Norm. Constant ($M$)}}
  & Ours w/ $M=512$ & 0.53 & -- \\
  & Ours w/ $M=1024$ & 0.53 & -- \\
  & Ours w/ $M=1536$ & 0.52 & -- \\

  \midrule

  \multirow{3}{*}{\centering \emph{Gaussian Noise ($\sigma_{\text{noise}}$)}}
  & $\sigma_{\text{noise}}=0$ & 0.50 & -- \\
  & $\sigma_{\text{noise}}=0.2$ & 0.53 & -- \\
  & $\sigma_{\text{noise}}=0.4$& 0.52 & -- \\

  \midrule

  \multirow{5}{*}{\centering \emph{RL Reward Components}}
  & SFT-only & 0.53 & -- \\
  & RL w/ DINOv2 only & 0.52 & -- \\
  & RL w/ SSIM only & 0.54 & -- \\
  & RL w/ LPIPS' only & 0.52 & -- \\
  & RL w/ DINOv2 + SSIM & 0.54 & -- \\
  & RL w/ DINOv2 + LPIPS' & 0.53 & -- \\
  & Full Reward & 0.54 & -- \\
  
  \midrule

  \multirow{3}{*}{\centering \emph{Precision Format}}
  & \texttt{float32} & 0.53 & -- \\
  & \texttt{bfloat16} & 0.52 & -- \\
  & \texttt{bfloat16} + \texttt{float32} & 0.53 & -- \\
  
  \bottomrule
 \end{tabular}
\end{table}

The \textit{Number Representation} provides an important insight. 
The superior performance of our continuous model over discrete baselines validates the core contribution of CNM. 
Next, with the \textit{Encoding Mechanism} we see a performance drop when replacing the \subcomponent{Fourier Feature Mapping}, which confirms its value in constructing an expressive \component{Number Encoder}. 
The \textit{Norm. Constant (M)} study shows the model's reasonable sensitivity to the canvas size, and justifies our choice of taking on a smaller value in $512$.
The \textit{RL Reward Components} results demonstrate the importance of a composite perceptual reward. 
Interestingly, we see that the biggest gain with respect to our reward composition comes not in SSIM but in DINOv2 performance (see Table~\ref{tab:main_results}).
While some individual metrics improve over the SFT-only baseline, the best results are achieved when all three complementary metrics are combined.
Moreover, combining all three discourages over-optimization along any axis.
For \textit{Gaussian Noise}, we find the best performance with moderate noise levels. 
\textit{Precision Format} shows that we get the best compute trade-off with \texttt{bfloat16} + \texttt{float32}, i.e. the \component{Number Encoder} and \component{Number Decoder} are in \texttt{float32} and everything else is \texttt{bfloat16}.
 
\subsection{Qualitative Analysis}
\label{subsec:qual_analysis}

Figure~\ref{fig:qualitative_results} presents a qualitative comparison for a representative SVG-Stack example.
The ground-truth image contains thin strokes and subtle positional differences that challenge both structural and numerical prediction.
StarVector introduces structural artifacts and misaligned segments.
OmniSVG fails to reproduce key foreground strokes.
Our model reconstructs stroke geometry and relative placement with higher fidelity, reflecting improved handling of local detail and spatial structure enabled by Continuous Number Modeling and perceptual fine-tuning.
\section{Related Work}
\label{sec:related}

Generative models for vector graphics have evolved from VAE-based approaches \cite{lopes2019learnedrepresentationscalablevectordiffsvg, carlier2020deepsvghierarchicalgenerativenetwork} to autoregressive transformers \cite{Rodriguez_2025_CVPR, xing2024llm4svg, Wang_2025svgen, wu2025chat2svgvectorgraphicsgeneration}.
However, prior autoregressive methods rely on discrete tokenization or quantization \cite{yang2025omnisvg}, which we have shown leads to inefficiency and loss of precision.
While some general-purpose methods treat numbers as continuous \cite{golkar2024xvalcontinuousnumericaltokenization}, they typically use simple linear projections that lack the expressivity needed for dense geometric data.
Our Continuous Number Modeling offers a distinct alternative: treating numerical values as first-class continuous variables encoded via Fourier features.
Furthermore, while reinforcement learning has been successfully applied to discrete SVG domains \cite{rodriguez2025renderingawarereinforcementlearningvector}, we show that this alignment paradigm yields even greater benefits when applied to continuous numerical representations.
We provide a more comprehensive review of related work in Appendix~\ref{app:related_work}.

\section{Conclusion}
\label{sec:conclusion}

We presented {Continuous Number Modeling (CNM)}, a framework for vector graphic generation that replaces digit-level tokenization of continuous values with a direct continuous representation. 
This formulation not only reduces sequence length and removes fragmentation effects, but also establishes a clearer correspondence between the numerical structure of an SVG and the model’s internal representation. 
The resulting improvement in geometric precision is amplified by an RL stage that optimizes a composite perceptual reward signal. 
Together, these components produce consistent gains in perceptual fidelity, structural accuracy, and inference efficiency on SVG-Stack.

\bibliography{main}

@string(CVPR= {IEEE Conf. Comput. Vis. Pattern Recog.})

@string(ICCV= {Int. Conf. Comput. Vis.})

@string(CVPR  = {CVPR})

@string(ICCV  = {ICCV})

@article{Gudeman1993RepresentingTI,
    title        = {Representing Type Information in Dynamically Typed Languages},
    author       = {David A. Gudeman},
    year         = {1993},
    journal      = {IEEE Transactions on Reliability},
    url          = {https://api.semanticscholar.org/CorpusID:18284374}
}

@article{1284395ssim,
    title        = {Image quality assessment: from error visibility to structural similarity},
    author       = {Zhou Wang and Bovik, A.C. and Sheikh, H.R. and Simoncelli, E.P.},
    year         = {2004},
    journal      = {IEEE Transactions on Image Processing},
    volume       = {13},
    number       = {4},
    pages        = {600--612},
    doi          = {10.1109/TIP.2003.819861}
}

@inproceedings{oord2016wavenetgenerativemodelraw,
    title        = {{WaveNet}: A Generative Model for Raw Audio},
    author       = {van den Oord, Aaron and Dieleman, Sander and Zen, Heiga and Simonyan, Karen and Vinyals, Oriol and Graves, Alex and Kalchbrenner, Nal and Senior, Andrew and Kavukcuoglu, Koray},
    year         = {2016},
    booktitle    = {9th ISCA Speech Synthesis Workshop},
    pages        = {125--125}
}

@inproceedings{sennrich2016neuralmachinetranslationrarbpe,
    title        = {Neural Machine Translation of Rare Words with Subword Units},
    author       = {Sennrich, Rico  and Haddow, Barry  and Birch, Alexandra},
    year         = {2016},
    month        = aug,
    booktitle    = {Proceedings of the 54th Annual Meeting of the Association for Computational Linguistics (Volume 1: Long Papers)},
    pages        = {1715--1725}
}

@misc{wu2016googlesneuralmachinetranslationwordpiece,
    title        = {Google's Neural Machine Translation System: Bridging the Gap between Human and Machine Translation},
    author       = {Yonghui Wu and Mike Schuster and Zhifeng Chen and Quoc V. Le and Mohammad Norouzi and Wolfgang Macherey and Maxim Krikun and Yuan Cao and Qin Gao and Klaus Macherey and Jeff Klingner and Apurva Shah and Melvin Johnson and Xiaobing Liu and Łukasz Kaiser and Stephan Gouws and Yoshikiyo Kato and Taku Kudo and Hideto Kazawa and Keith Stevens and George Kurian and Nishant Patil and Wei Wang and Cliff Young and Jason Smith and Jason Riesa and Alex Rudnick and Oriol Vinyals and Greg Corrado and Macduff Hughes and Jeffrey Dean},
    year         = {2016},
    url          = {https://arxiv.org/abs/1609.08144},
    eprint       = {1609.08144},
    archiveprefix = {arXiv},
    primaryclass = {cs.CL}
}

@inproceedings{oord2018neuraldiscreterepresentationlearningvqvae,
    title        = {Neural Discrete Representation Learning},
    author       = {van den Oord, Aaron and Vinyals, Oriol and Kavukcuoglu, Koray},
    year         = {2017},
    booktitle    = {Advances in Neural Information Processing Systems},
    volume       = {30}
}

@inproceedings{zhang2018unreasonableeffectivenessdeepfeatureslpips,
    title        = {The Unreasonable Effectiveness of Deep Features as a Perceptual Metric},
    author       = {Zhang, Richard and Isola, Phillip and Efros, Alexei A. and Shechtman, Eli and Wang, Oliver},
    year         = {2018},
    month        = {June},
    booktitle    = {Proceedings of the IEEE Conference on Computer Vision and Pattern Recognition (CVPR)}
}

@inproceedings{lopes2019learnedrepresentationscalablevectordiffsvg,
    title        = {A Learned Representation for Scalable Vector Graphics},
    author       = {Lopes, Raphael Gontijo and Ha, David and Eck, Douglas and Shlens, Jonathon},
    year         = {2019},
    month        = {October},
    booktitle    = {Proceedings of the IEEE/CVF International Conference on Computer Vision (ICCV)}
}

@article{Radford2019LanguageMAgpt2,
    title        = {Language Models are Unsupervised Multitask Learners},
    author       = {Radford, Alec and Wu, Jeffrey and Child, Rewon and Luan, David and Amodei, Dario and Sutskever, Ilya},
    year         = {2019},
    journal      = {OpenAI},
    url          = {https://cdn.openai.com/better-language-models/language_models_are_unsupervised_multitask_learners.pdf},
    note         = {Accessed: 2026-01-28}
}

@inproceedings{carlier2020deepsvghierarchicalgenerativenetwork,
    title        = {{DeepSVG}: A Hierarchical Generative Network for Vector Graphics Animation},
    author       = {Carlier, Alexandre and Danelljan, Martin and Alahi, Alexandre and Timofte, Radu},
    year         = {2020},
    booktitle    = {Advances in Neural Information Processing Systems},
    volume       = {33}
}

@inproceedings{ouyang2022traininglanguagemodelsfollowrlhf,
    title        = {Training Language Models to Follow Instructions with Human Feedback},
    author       = {Ouyang, Long and Wu, Jeffrey and Jiang, Xu and Almeida, Diogo and Wainwright, Carroll and Mishkin, Pamela and Zhang, Chong and Agarwal, Sandhini and Slama, Katarina and Ray, Alex and Schulman, John and Hilton, Jacob and Kelton, Fraser and Miller, Luke and Simens, Maddie and Askell, Amanda and Welinder, Peter and Christiano, Paul F and Leike, Jan and Lowe, Ryan},
    year         = {2022},
    booktitle    = {Advances in Neural Information Processing Systems},
    volume       = {35}
}

@inproceedings{jain2022vectorfusiontexttosvgabstractingpixelbased,
    title        = {{VectorFusion}: Text-to-{SVG} by Abstracting Pixel-Based Diffusion Models},
    author       = {Jain, Ajay and Xie, Amber and Abbeel, Pieter},
    year         = {2023},
    month        = {June},
    booktitle    = {Proceedings of the IEEE/CVF Conference on Computer Vision and Pattern Recognition (CVPR)},
    pages        = {19197--19206}
}

@misc{lee2023aligningtexttoimagemodelsusingrlhf,
    title        = {Aligning Text-to-Image Models using Human Feedback},
    author       = {Kimin Lee and Hao Liu and Moonkyung Ryu and Olivia Watkins and Yuqing Du and Craig Boutilier and Pieter Abbeel and Mohammad Ghavamzadeh and Shixiang Shane Gu},
    year         = {2023},
    url          = {https://arxiv.org/abs/2302.12192},
    eprint       = {2302.12192},
    archiveprefix = {arXiv},
    primaryclass = {cs.LG}
}

@inproceedings{black2024trainingdiffusionmodelsreinforcement,
    title        = {Training Diffusion Models with Reinforcement Learning},
    author       = {Kevin Black and Michael Janner and Yilun Du and Ilya Kostrikov and Sergey Levine},
    year         = {2024},
    booktitle    = {The Twelfth International Conference on Learning Representations}
}

@misc{golkar2024xvalcontinuousnumericaltokenization,
    title        = {xVal: A Continuous Numerical Tokenization for Scientific Language Models},
    author       = {Siavash Golkar and Mariel Pettee and Michael Eickenberg and Alberto Bietti and Miles Cranmer and Geraud Krawezik and Francois Lanusse and Michael McCabe and Ruben Ohana and Liam Parker and Bruno Régaldo-Saint Blancard and Tiberiu Tesileanu and Kyunghyun Cho and Shirley Ho},
    year         = {2024},
    url          = {https://arxiv.org/abs/2310.02989},
    eprint       = {2310.02989},
    archiveprefix = {arXiv},
    primaryclass = {stat.ML}
}

@misc{ogezi2024optimizingnegativepromptsenhancednegopt,
    title        = {Optimizing Negative Prompts for Enhanced Aesthetics and Fidelity in Text-To-Image Generation},
    author       = {Michael Ogezi and Ning Shi},
    year         = {2024},
    url          = {https://arxiv.org/abs/2403.07605},
    eprint       = {2403.07605},
    archiveprefix = {arXiv},
    primaryclass = {cs.CV}
}

@article{oquab2024dinov2learningrobustvisual,
    title        = {{DINOv2}: Learning Robust Visual Features without Supervision},
    author       = {Oquab, Maxime and Darcet, Timothée and Moutakanni, Théo and Vo, Huy and Szafraniec, Marc and Khalidov, Vasil and Fernandez, Pierre and Haziza, Daniel and Massa, Francisco and El-Nouby, Alaaeldin and Assran, Mahmoud and Ballas, Nicolas and Galuba, Wojciech and Howes, Russell and Huang, Po-Yao and Li, Shang-Wen and Misra, Ishan and Rabbat, Michael and Sharma, Vasu and Synnaeve, Gabriel and Xu, Hu and Jegou, Hervé and Mairal, Julien and Labatut, Patrick and Joulin, Armand and Bojanowski, Piotr},
    year         = {2024},
    journal      = {Transactions on Machine Learning Research},
    issn         = {2835-8856},
    url          = {https://openreview.net/forum?id=a68SUt6zLL}
}

@misc{shao2024deepseekmathpushinglimitsmathematicalgrpo,
    title        = {DeepSeekMath: Pushing the Limits of Mathematical Reasoning in Open Language Models},
    author       = {Zhihong Shao and Peiyi Wang and Qihao Zhu and Runxin Xu and Junxiao Song and Xiao Bi and Haowei Zhang and Mingchuan Zhang and Y. K. Li and Y. Wu and Daya Guo},
    year         = {2024},
    url          = {https://arxiv.org/abs/2402.03300},
    eprint       = {2402.03300},
    archiveprefix = {arXiv},
    primaryclass = {cs.CL}
}

@misc{wang2024qwen2vlenhancingvisionlanguagemodels,
    title        = {Qwen2-VL: Enhancing Vision-Language Model's Perception of the World at Any Resolution},
    author       = {Peng Wang and Shuai Bai and Sinan Tan and Shijie Wang and Zhihao Fan and Jinze Bai and Keqin Chen and Xuejing Liu and Jialin Wang and Wenbin Ge and Yang Fan and Kai Dang and Mengfei Du and Xuancheng Ren and Rui Men and Dayiheng Liu and Chang Zhou and Jingren Zhou and Junyang Lin},
    year         = {2024},
    url          = {https://arxiv.org/abs/2409.12191},
    eprint       = {2409.12191},
    archiveprefix = {arXiv},
    primaryclass = {cs.CV}
}

@article{xing2024llm4svg,
    title        = {Empowering LLMs to Understand and Generate Complex Vector Graphics},
    author       = {Xing, Ximing and Hu, Juncheng and Liang, Guotao and Zhang, Jing and Xu, Dong and Yu, Qian},
    year         = {2024},
    url          = {https://arxiv.org/abs/2412.11102}
}

@misc{qwen2025qwen25technicalreport,
    title        = {Qwen2.5 Technical Report},
    author       = {An Yang and Baosong Yang and Beichen Zhang and Binyuan Hui and Bo Zheng and Bowen Yu and Chengyuan Li and Dayiheng Liu and Fei Huang and Haoran Wei and Huan Lin and Jian Yang and Jianhong Tu and Jianwei Zhang and Jianxin Yang and Jiaxi Yang and Jingren Zhou and Junyang Lin and Kai Dang and Keming Lu and Keqin Bao and Kexin Yang and Le Yu and Mei Li and Mingfeng Xue and Pei Zhang and Qin Zhu and Rui Men and Runji Lin and Tianhao Li and Tianyi Tang and Tingyu Xia and Xingzhang Ren and Xuancheng Ren and Yang Fan and Yang Su and Yichang Zhang and Yu Wan and Yuqiong Liu and Zeyu Cui and Zhenru Zhang and Zihan Qiu},
    year         = {2025},
    url          = {https://arxiv.org/abs/2412.15115},
    eprint       = {2412.15115},
    archiveprefix = {arXiv},
    primaryclass = {cs.CL}
}

@inproceedings{Rodriguez_2025_CVPR,
    title        = {StarVector: Generating Scalable Vector Graphics Code from Images and Text},
    author       = {Rodriguez, Juan A. and Puri, Abhay and Agarwal, Shubham and Laradji, Issam H. and Rodriguez, Pau and Rajeswar, Sai and Vazquez, David and Pal, Christopher and Pedersoli, Marco},
    year         = {2025},
    month        = {June},
    booktitle    = {Proceedings of the IEEE/CVF Conference on Computer Vision and Pattern Recognition (CVPR)},
    pages        = {16175--16186}
}

@inproceedings{rodriguez2025renderingawarereinforcementlearningvector,
    title        = {Rendering-Aware Reinforcement Learning for Vector Graphics Generation},
    author       = {Juan A. Rodriguez and Haotian Zhang and Abhay Puri and Aarash Feizi and Rishav Pramanik and Pascal Wichmann and Arnab Mondal and Mohammad Reza Samsami and Rabiul Awal and Perouz Taslakian and Spandana Gella and Sai Rajeswar and David Vazquez and Christopher Pal and Marco Pedersoli},
    year         = {2025},
    booktitle    = {Advances in Neural Information Processing Systems}
}

@inproceedings{Wang_2025svgen,
    title        = {SVGen: Interpretable Vector Graphics Generation with Large Language Models},
    author       = {Wang, Feiyu and Zhao, Zhiyuan and Liu, Yuandong and Zhang, Da and Gao, Junyu and Sun, Hao and Li, Xuelong},
    year         = {2025},
    month        = oct,
    booktitle    = {Proceedings of the 33rd ACM International Conference on Multimedia},
    publisher    = {ACM},
    series       = {MM ’25},
    pages        = {9608–9617},
    doi          = {10.1145/3746027.3755011},
    url          = {http://dx.doi.org/10.1145/3746027.3755011},
    collection   = {MM ’25}
}

@misc{wu2025chat2svgvectorgraphicsgeneration,
    title        = {Chat2SVG: Vector Graphics Generation with Large Language Models and Image Diffusion Models},
    author       = {Ronghuan Wu and Wanchao Su and Jing Liao},
    year         = {2025},
    url          = {https://arxiv.org/abs/2411.16602},
    eprint       = {2411.16602},
    archiveprefix = {arXiv},
    primaryclass = {cs.CV}
}

@inproceedings{yang2025omnisvg,
    title        = {Omnisvg: A unified scalable vector graphics generation model},
    author       = {Yang, Yiying and Cheng, Wei and Chen, Sijin and Zeng, Xianfang and Yin, Fukun and Zhang, Jiaxu and Wang, Liao and Yu, Gang and Ma, Xingjun and Jiang, Yu-Gang},
    year         = {2025},
    booktitle    = {Advances in Neural Information Processing Systems}
}

@misc{yang2025qwen3technicalreport,
    title        = {Qwen3 Technical Report},
    author       = {An Yang and Anfeng Li and Baosong Yang and Beichen Zhang and Binyuan Hui and Bo Zheng and Bowen Yu and Chang Gao and Chengen Huang and Chenxu Lv and Chujie Zheng and Dayiheng Liu and Fan Zhou and Fei Huang and Feng Hu and Hao Ge and Haoran Wei and Huan Lin and Jialong Tang and Jian Yang and Jianhong Tu and Jianwei Zhang and Jianxin Yang and Jiaxi Yang and Jing Zhou and Jingren Zhou and Junyang Lin and Kai Dang and Keqin Bao and Kexin Yang and Le Yu and Lianghao Deng and Mei Li and Mingfeng Xue and Mingze Li and Pei Zhang and Peng Wang and Qin Zhu and Rui Men and Ruize Gao and Shixuan Liu and Shuang Luo and Tianhao Li and Tianyi Tang and Wenbiao Yin and Xingzhang Ren and Xinyu Wang and Xinyu Zhang and Xuancheng Ren and Yang Fan and Yang Su and Yichang Zhang and Yinger Zhang and Yu Wan and Yuqiong Liu and Zekun Wang and Zeyu Cui and Zhenru Zhang and Zhipeng Zhou and Zihan Qiu},
    year         = {2025},
    url          = {https://arxiv.org/abs/2505.09388},
    eprint       = {2505.09388},
    archiveprefix = {arXiv},
    primaryclass = {cs.CL}
}
\bibliographystyle{icml2025}

\newpage
\appendix
\onecolumn

\section{Broader Related Work}
\label{app:related_work}

Our work lies at the intersection of three research areas: generative modeling for vector graphics, numerical representation in sequence models, and reinforcement learning for perceptual alignment.

\paragraph{Generative models for vector graphics.}

Early approaches to vector graphic generation, including SVG-VAE \cite{lopes2019learnedrepresentationscalablevectordiffsvg} and DeepSVG \cite{carlier2020deepsvghierarchicalgenerativenetwork}, relied on variational autoencoders. 
These models performed well for simple icons and characters but were unable to model complex, multi-part structures. 
The field then shifted toward autoregressive transformers that treat SVG synthesis as sequence modeling, exemplified by StarVector \cite{Rodriguez_2025_CVPR}, LLM4SVG \cite{xing2024llm4svg}, SVGen \cite{Wang_2025svgen}, and Chat2SVG \cite{wu2025chat2svgvectorgraphicsgeneration}.
These methods capture compositional structure more effectively but still inherit inefficiencies from discrete token representations of continuous coordinates. 
OmniSVG \cite{yang2025omnisvg} attempts to mitigate this by quantizing coordinates to reduce sequence length, which enables longer and more complex SVGs but introduces quantization artifacts, limits numerical precision, and does not preserve quantitative relationships between bins.

A parallel line of research explored diffusion-based generation.
Methods such as VectorFusion \cite{jain2022vectorfusiontexttosvgabstractingpixelbased} synthesize SVG primitives with a diffusion process and rely on differentiable rasterizers to provide pixel-domain supervision.

Although the most powerful of the above approaches have improved vector graphic generation considerably, they rely on discrete encoding of continuous parameters, either through vanilla tokenization or quantization. 
Our method removes this limitation by representing numeric values directly as continuous variables.

\paragraph{Number representation in generative models.}

The difficulty of representing continuous numerical data extends beyond vector graphics. 
In natural language processing, standard subword tokenization schemes such as BPE \cite{sennrich2016neuralmachinetranslationrarbpe} and WordPiece \cite{wu2016googlesneuralmachinetranslationwordpiece} fragment floating-point values like \texttt{528.491} into multiple tokens. 
This expansion increases sequence length and forces models to learn numeric syntax that is orthogonal to the task.
Some models, including GPT-2 \cite{Radford2019LanguageMAgpt2}, decrease fragmentation by hard-coding a subset of integers into the vocabulary,\footnote{\url{https://platform.openai.com/tokenizer}} which reduces the depth of decomposition but still introduces discretization artifacts and inherits the limitations observed in OmniSVG.

Quantization offers an alternative, by mapping continuous values to a finite codebook or bin. 
This approach, used in WaveNet \citep[for audio generation]{oord2016wavenetgenerativemodelraw}, VQ-VAE \citep[for representation learning]{oord2018neuraldiscreterepresentationlearningvqvae}, and most relevantly OmniSVG \cite{yang2025omnisvg}, transforms regression into classification but introduces quantization error and limits precision.
Conversely, while xVal \cite{golkar2024xvalcontinuousnumericaltokenization} treats numbers as continuous for scientific tasks, its linear projection approach remains inadequate.

In contrast, our method treats numbers as first-class continuous variables, avoiding token fragmentation and quantization artifacts while making scale differences between numbers explicit and smooth.

\paragraph{Perceptual alignment with RL.}

Supervised learning with token- or pixel-level objectives does not always correlate with perceptual quality. 
Recent generative systems address this gap through RL fine-tuning to align outputs with preference-based or perceptual objectives. 
Reinforcement Learning from Human Feedback (RLHF) is the canonical example, used to align language and vision–language models with preference signals \cite{ouyang2022traininglanguagemodelsfollowrlhf}.
In image generation, RL-based alignment extends to human feedback \cite{lee2023aligningtexttoimagemodelsusingrlhf} and automated, non-differentiable metrics \cite{ogezi2024optimizingnegativepromptsenhancednegopt, black2024trainingdiffusionmodelsreinforcement}, with very recent work exploring RL for improving SVG rendering quality \cite{rodriguez2025renderingawarereinforcementlearningvector}.

Our work builds upon this alignment paradigm but applies it to a continuous action space. 
Unlike prior approaches that optimize discrete tokens, we use RL to fine-tune continuous numerical values directly, enabling precise geometric adjustments that maximize a composite perceptual reward based on SSIM \cite{1284395ssim}, LPIPS \cite{zhang2018unreasonableeffectivenessdeepfeatureslpips}, and DINOv2 \cite{oquab2024dinov2learningrobustvisual}.

\section{Additional Architectural Details}
\label{app:architecture}

\begin{figure}[!h]
	\centering
	\includegraphics[width=1.0\linewidth]{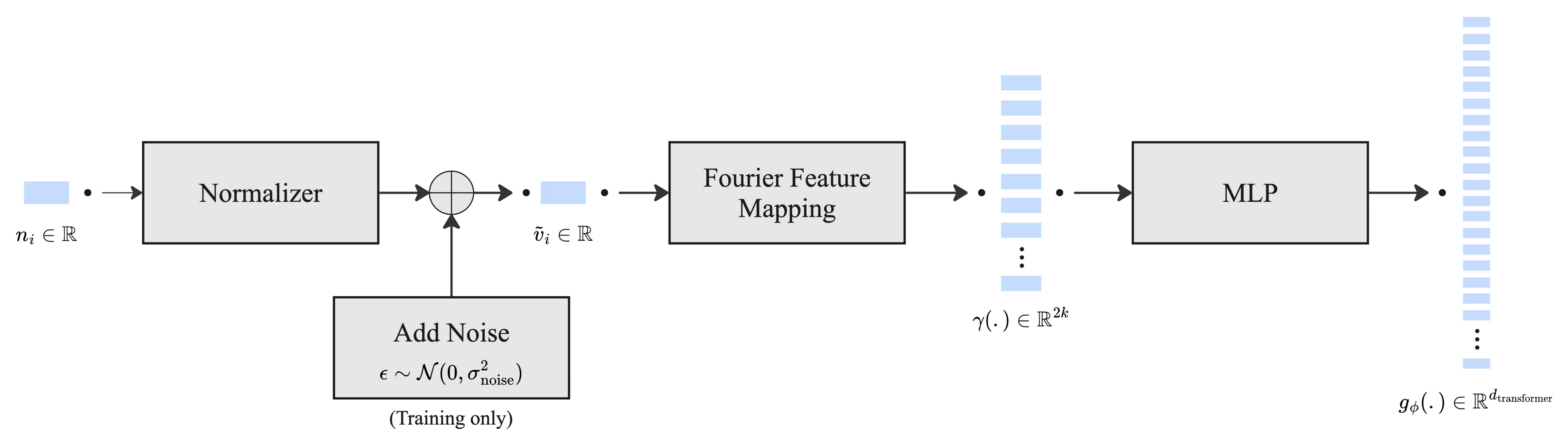}
	\caption{The \component{Number Encoder} architecture: 
    The encoder takes a raw number $n_i$, normalizes it to $v_i$, noises it to $\tilde{v}_i$ (if in training mode), applies the \subcomponent{Fourier Feature Mapping} with $\gamma(.)$, and projects to the transformer's input embedding space via the MLP with $g_\phi(.)$.
    The \component{Number Encoder}, in context, can be found in Figure~\ref{fig:training}, for training, and Figure~\ref{fig:inference}, for inference.
    }
	\label{fig:number_encoder}
\end{figure}

\begin{figure}[!h]
	\centering
	\includegraphics[width=0.70\linewidth]{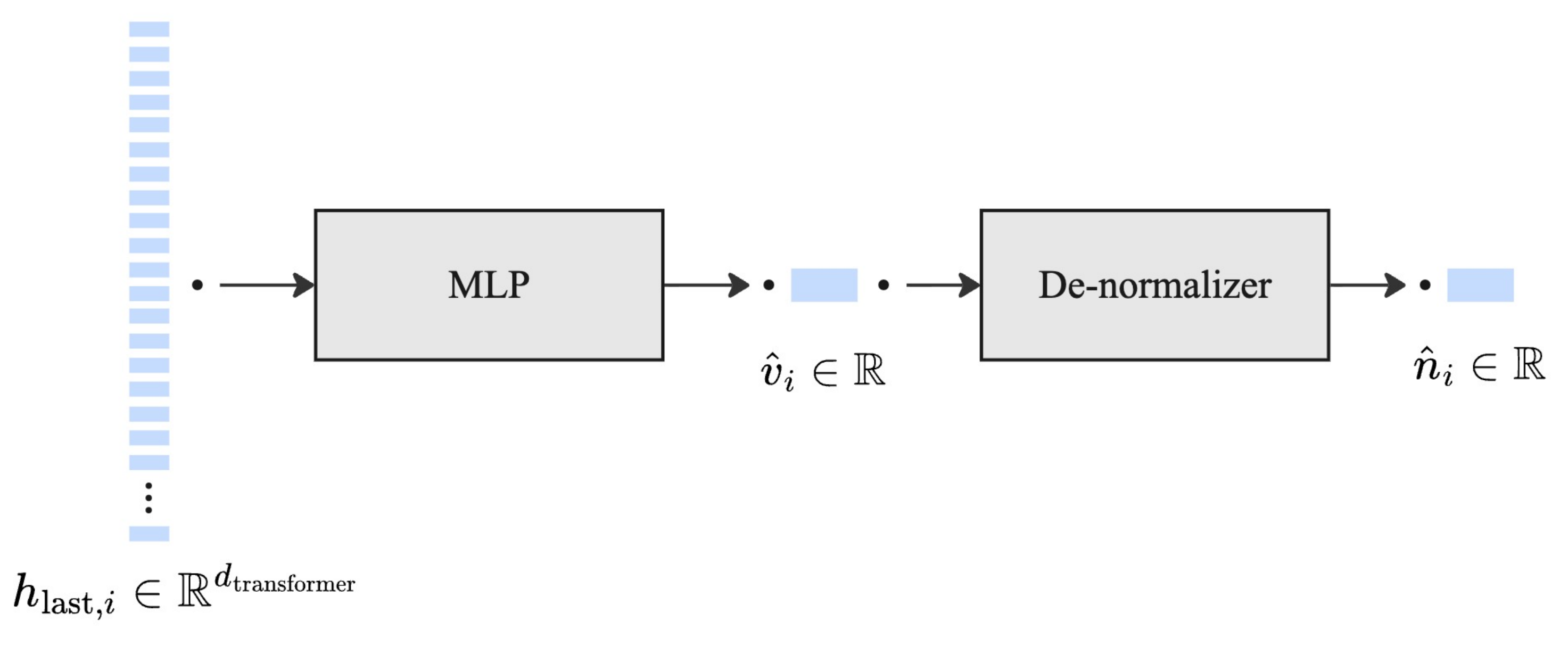}
	\caption{The \component{Number Decoder} architecture: 
    The decoder takes the last hidden state $h_{\text{last}, i}$ from the transformer, regresses it to a normalized scalar value $\hat{v}_i$ via an MLP with $f_\theta(.)$, and de-normalizes that to obtain the final prediction $\hat{n}_i$.
    The \component{Number Decoder}, in context, can be found in Figure~\ref{fig:training}, for training, and Figure~\ref{fig:inference}, for inference.
    }
	\label{fig:number_decoder}
\end{figure}

We provide diagrammatic expansions of the \component{Number Encoder} (Figure~\ref{fig:number_encoder}) and \component{Number Decoder} (Figure~\ref{fig:number_decoder}).

\section{RL Fine-Tuning}
\begin{figure*}[!h]
  \centering
  \includegraphics[width=\linewidth]{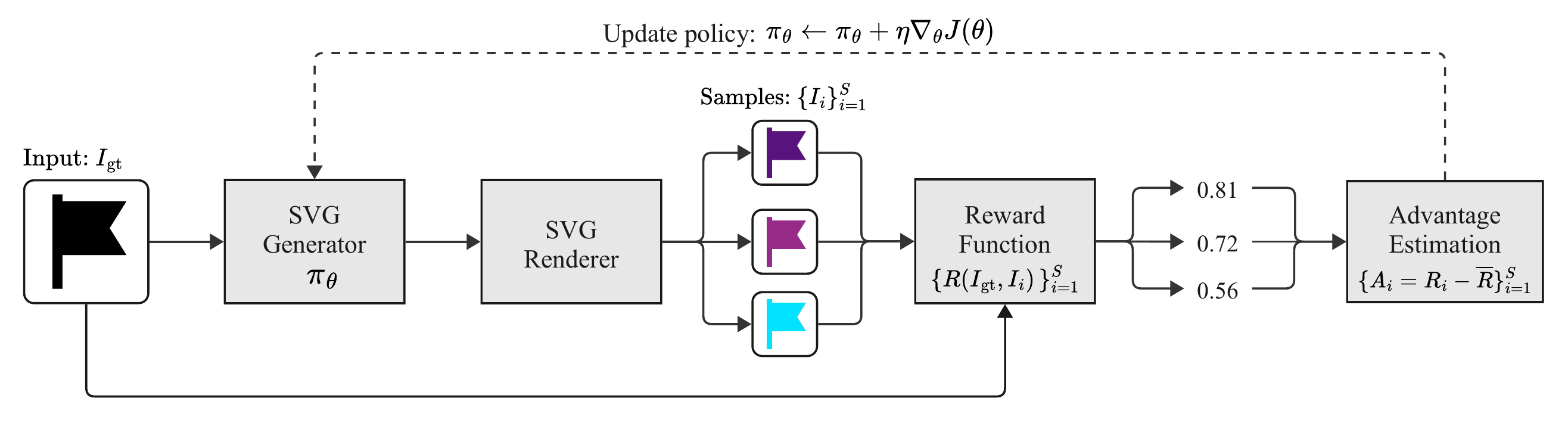}
  \caption{
    Overview of the RL fine-tuning stage using GRPO. 
    The generator policy $\pi_{\theta}$ produces multiple SVG samples $\{I_i\}_{i=1}^{S}$, which are rendered and compared against the ground-truth image $I_{\text{gt}}$ by a perceptual reward function $R(I_{\text{gt}}, I_i)$ (see Subsection \ref{subsec:reward_function}). 
    Continuous reward values guide advantage estimation $\{A_i = R_i - \overline{R}\}_{i=1}^{S}$, 
    and the policy is updated via gradient ascent to maximize perceptual fidelity.
}
  \label{fig:reinforcement-learning}
\end{figure*}

We provide an overview of our RL training process in Figure~\ref{fig:reinforcement-learning}.

\section{Additional Qualitative Results}
\label{app:qualitative}

\paragraph{Qualitative trends.} 
Across the examples in Table~\ref{tab:qualitative_results}, the top rows verify that all methods can handle simple glyph-style icons, though baselines often distort stroke width or drop negative space. As the examples progress toward more structured silhouettes, path-intensive icons, and cluttered multi-part compositions (bottom rows), competing methods increasingly oversimplify shapes, lose small sub-elements, or collapse into invalid outputs. Our method maintains both global layout and local structure closer to the ground truth.

\newlength{\imgwidth}
\setlength{\imgwidth}{0.19\linewidth}

{
\small
\setlength{\tabcolsep}{3pt} 
\begin{longtable}{ @{} m{\imgwidth} m{\imgwidth} m{\imgwidth} m{\imgwidth} 
@{} 
}

\caption{Additional qualitative samples. Empty squares denote invalid SVGs or fully transparent renders.}
\label{tab:qualitative_results} \\

\toprule
\centering \textbf{Ground Truth} &
\centering \textbf{Ours} &
\centering \textbf{StarVector} &
\centering\arraybackslash \textbf{OmniSVG}
\\
\midrule
\endfirsthead

\caption[]{Additional qualitative samples (continued).} \\
\toprule
\centering \textbf{Ground Truth} &
\centering \textbf{Ours} &
\centering \textbf{StarVector} &
\centering\arraybackslash \textbf{OmniSVG}
\\
\midrule
\endhead

\bottomrule
\endfoot


\includegraphics[width=\linewidth]{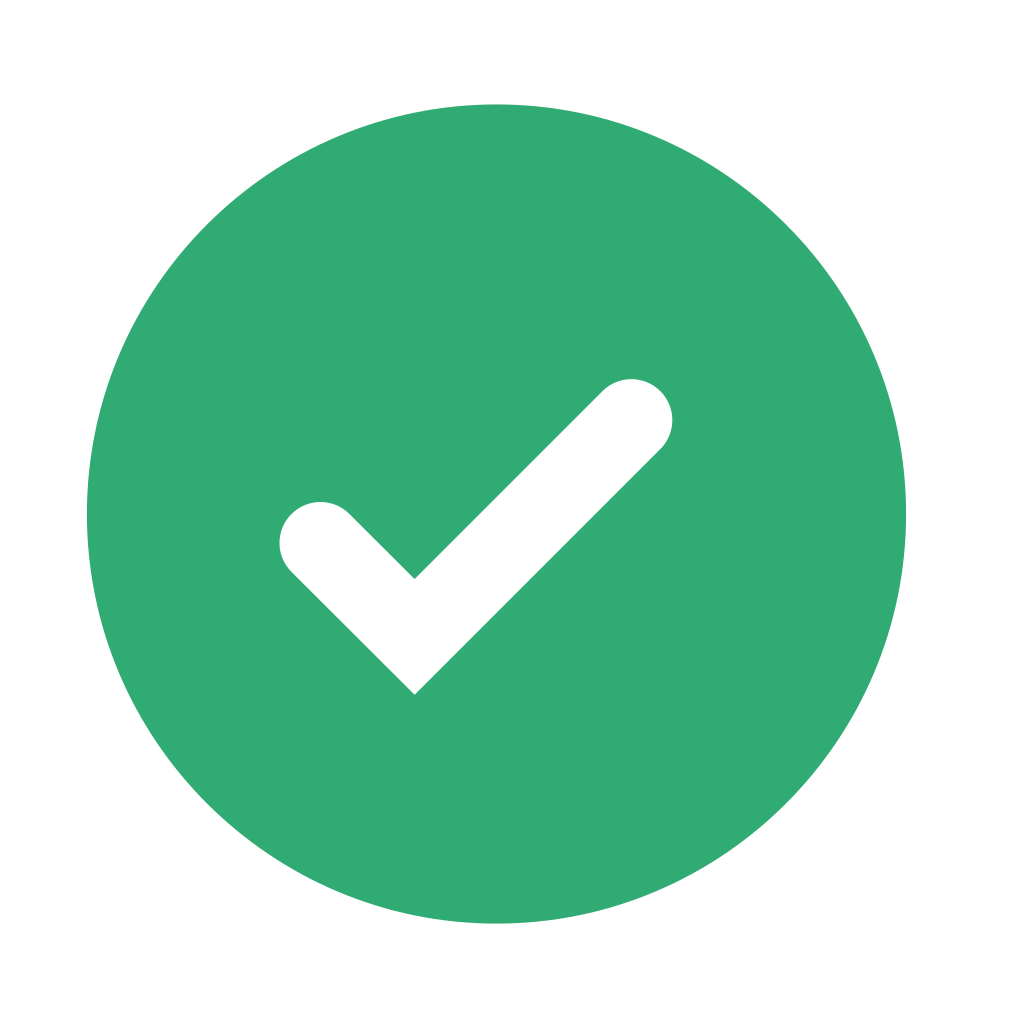} &
\includegraphics[width=\linewidth]{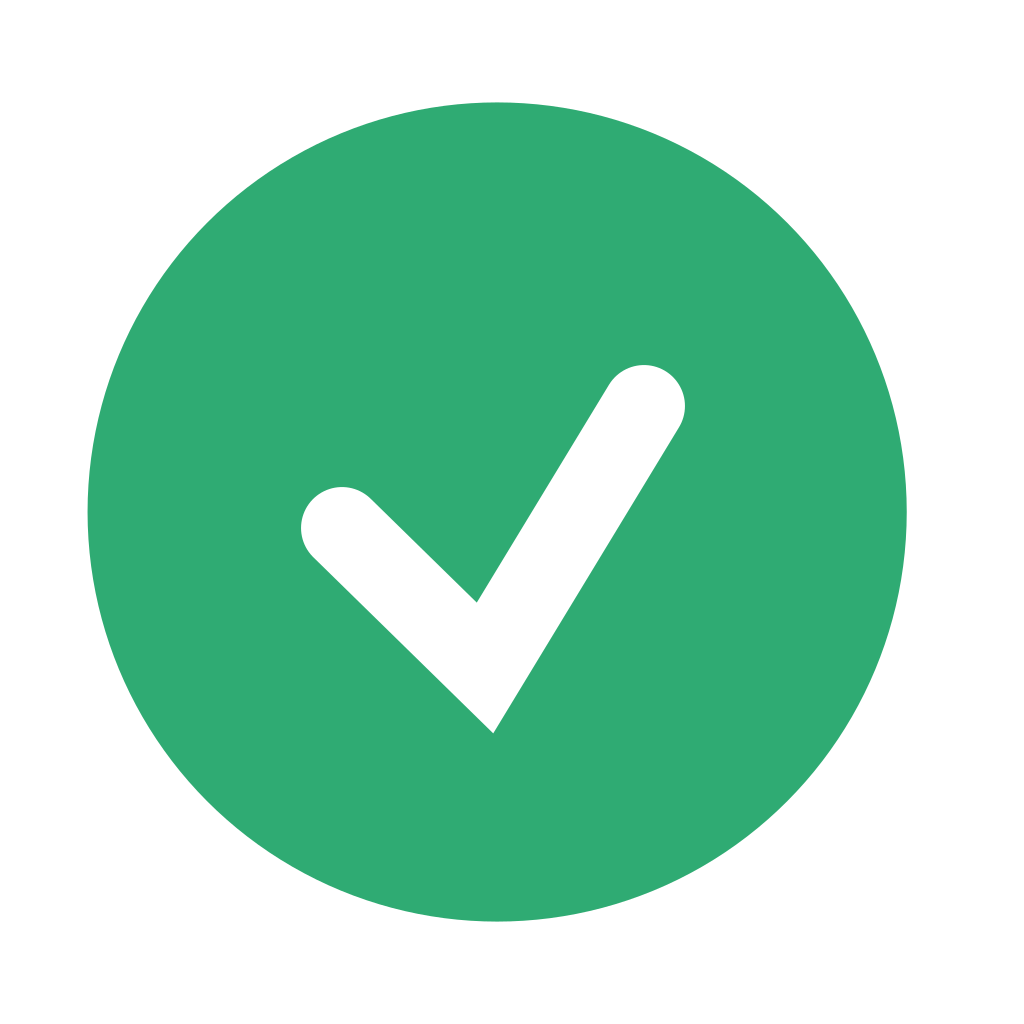} &
\includegraphics[width=\linewidth]{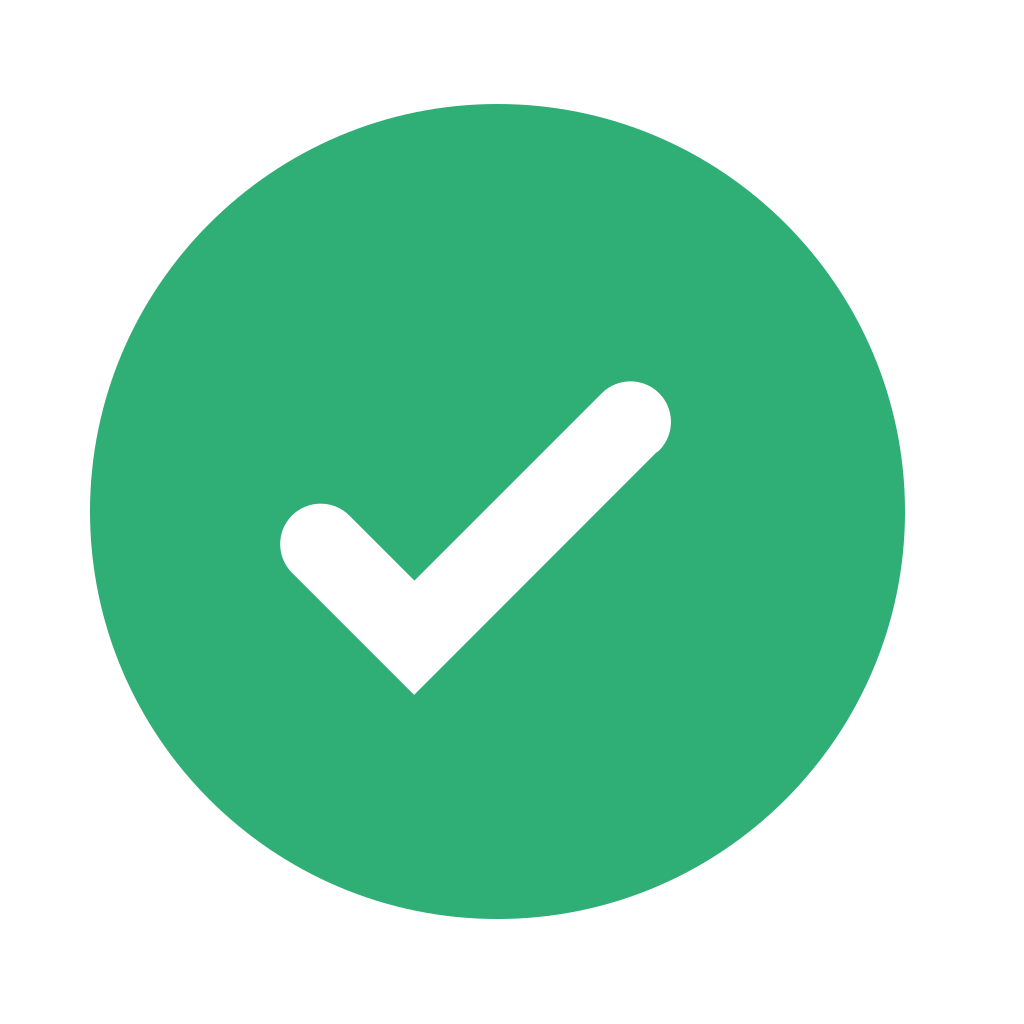} &
\includegraphics[width=\linewidth]{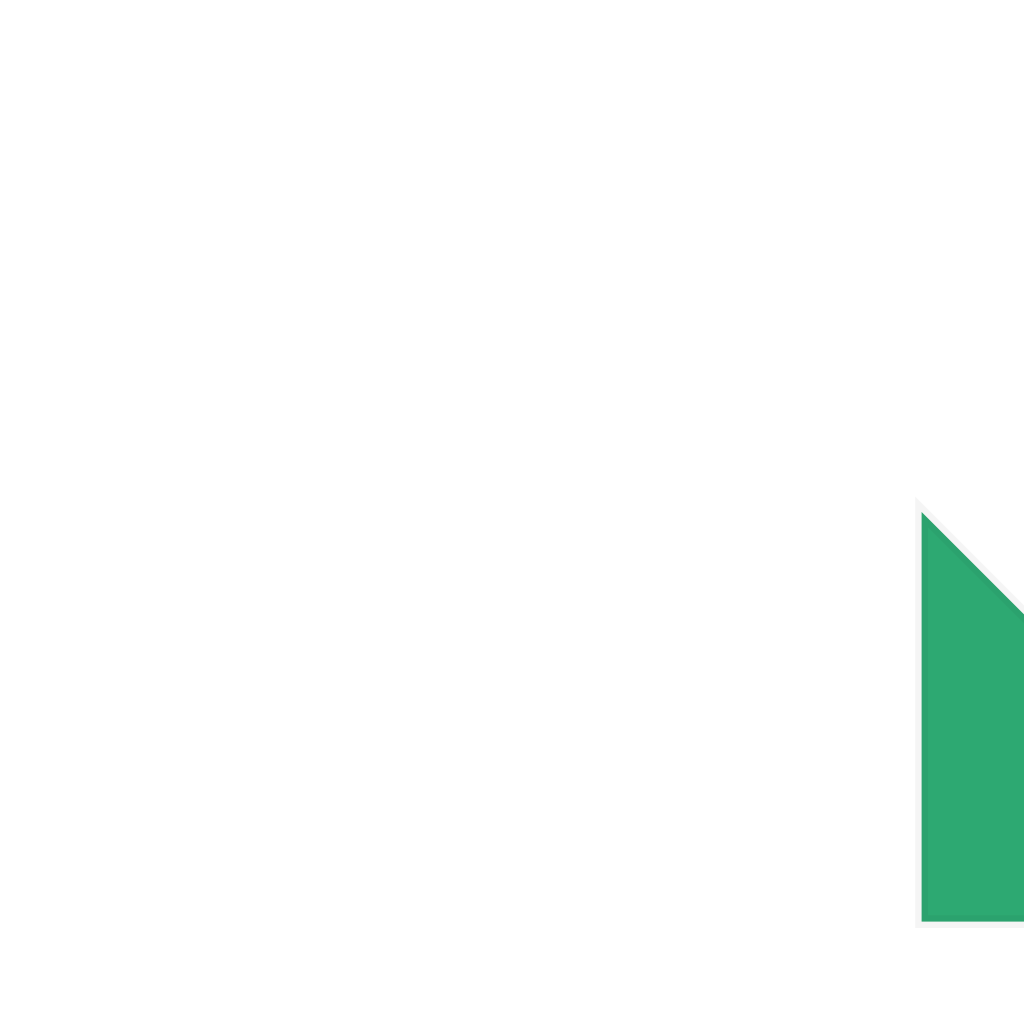} 
\\
\addlinespace[1.5em] 

\includegraphics[width=\linewidth]{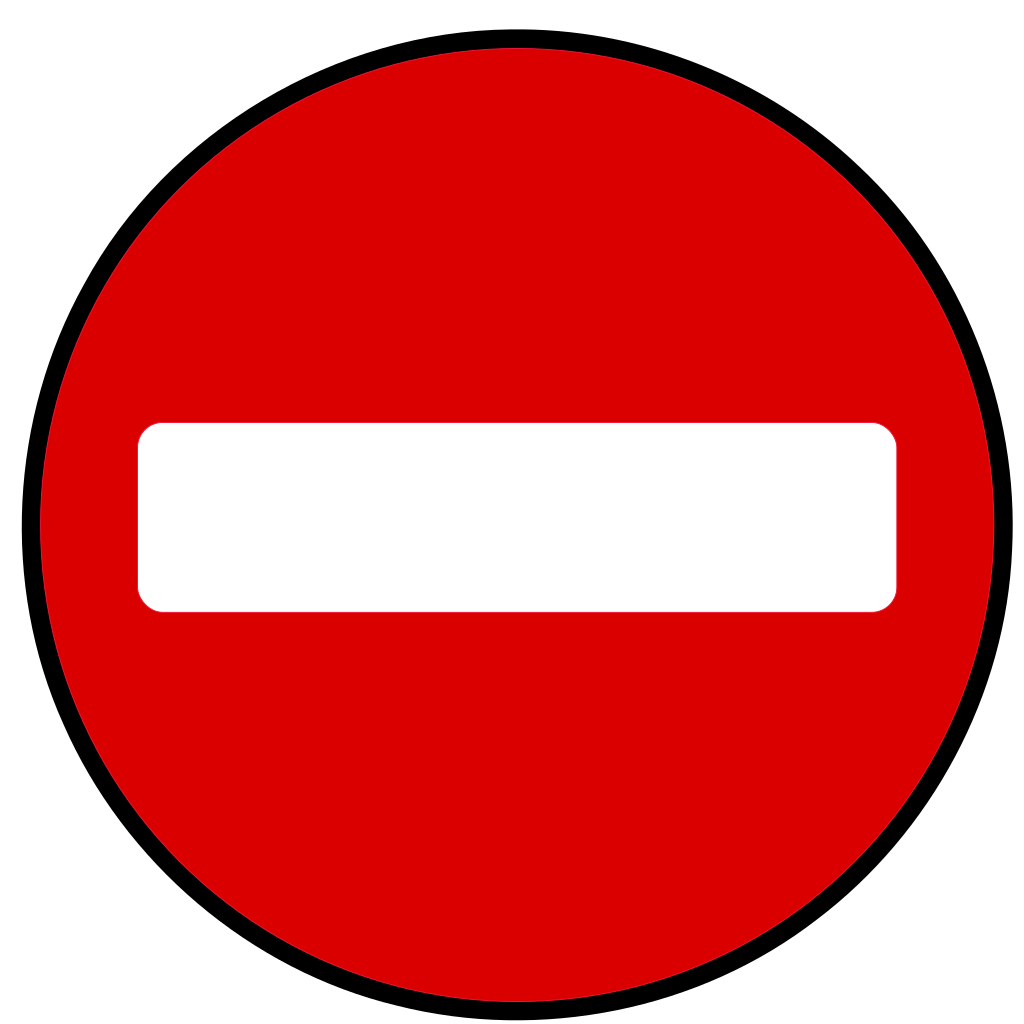} &
\includegraphics[width=\linewidth]{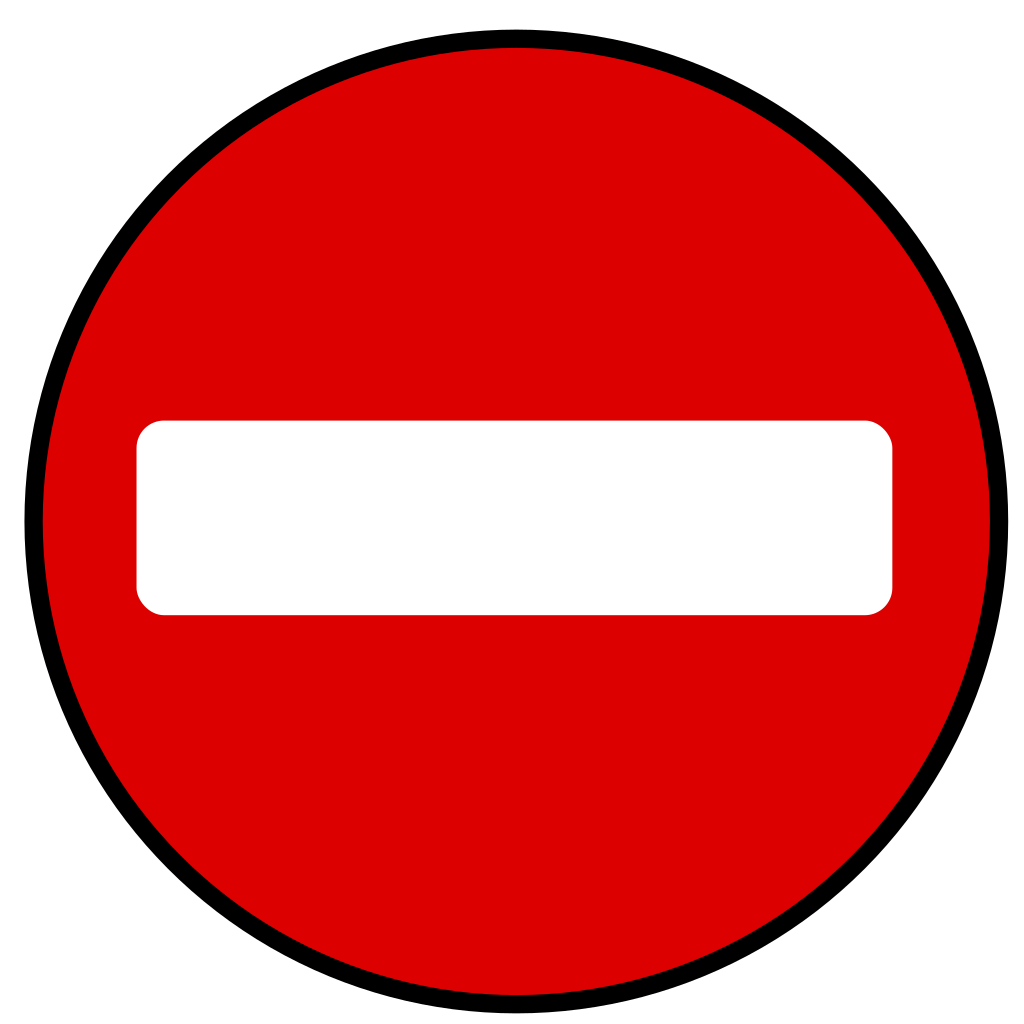} &
\includegraphics[width=\linewidth]{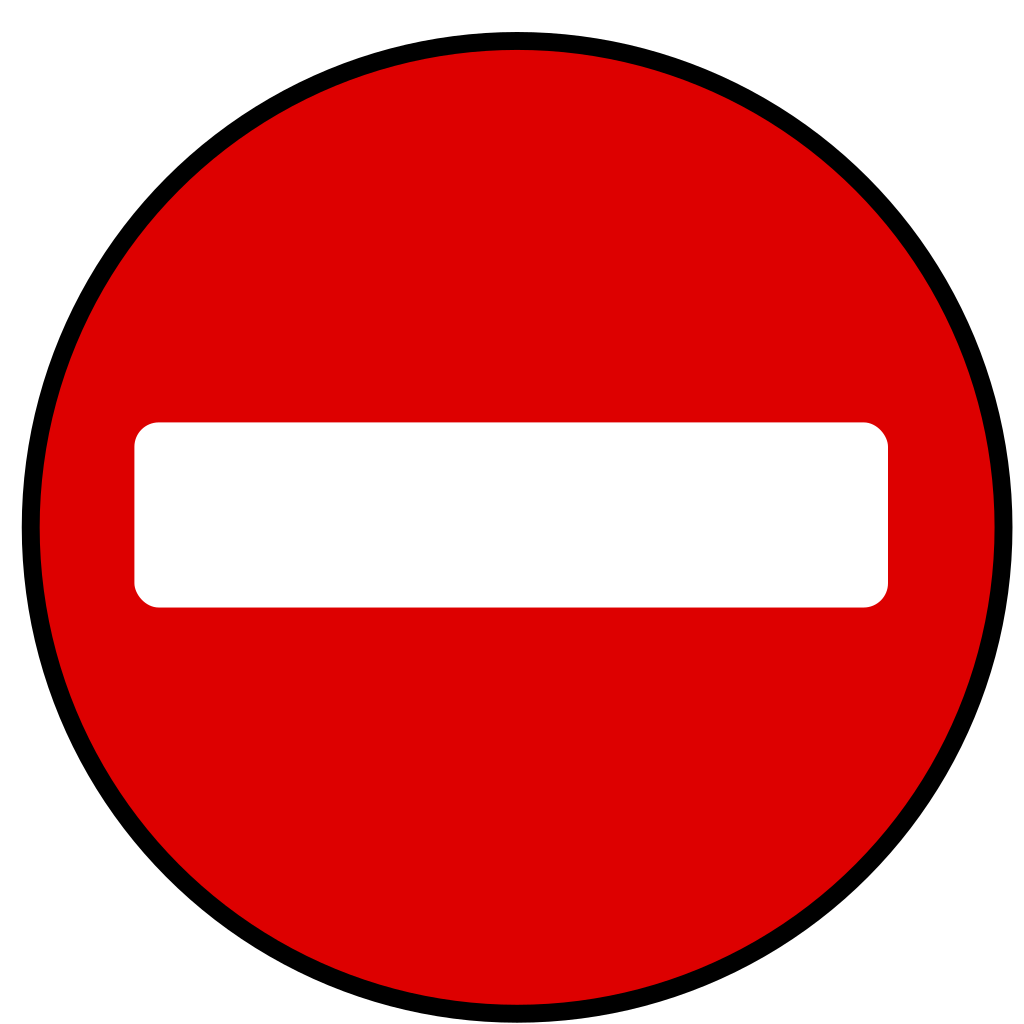} &
\includegraphics[width=\linewidth]{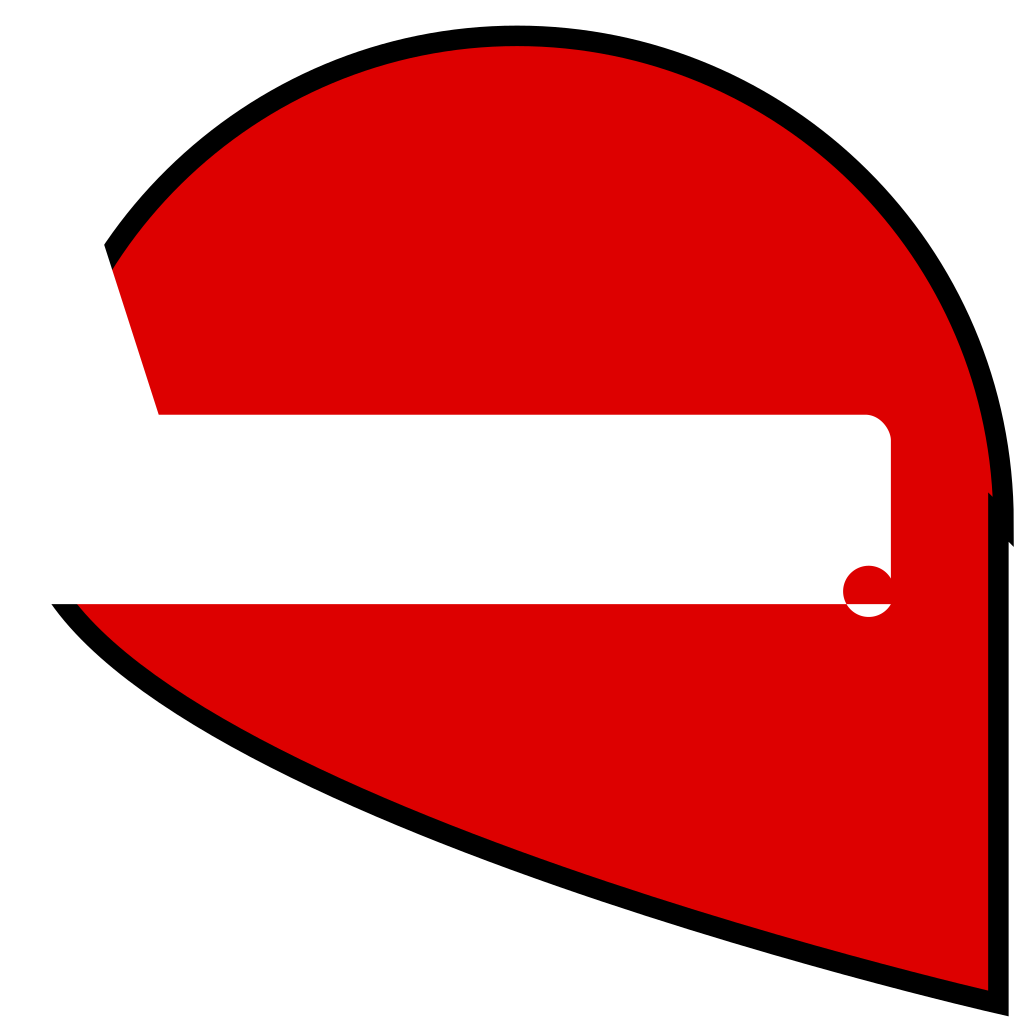} 
\\
\addlinespace[1.5em]

\includegraphics[width=\linewidth]{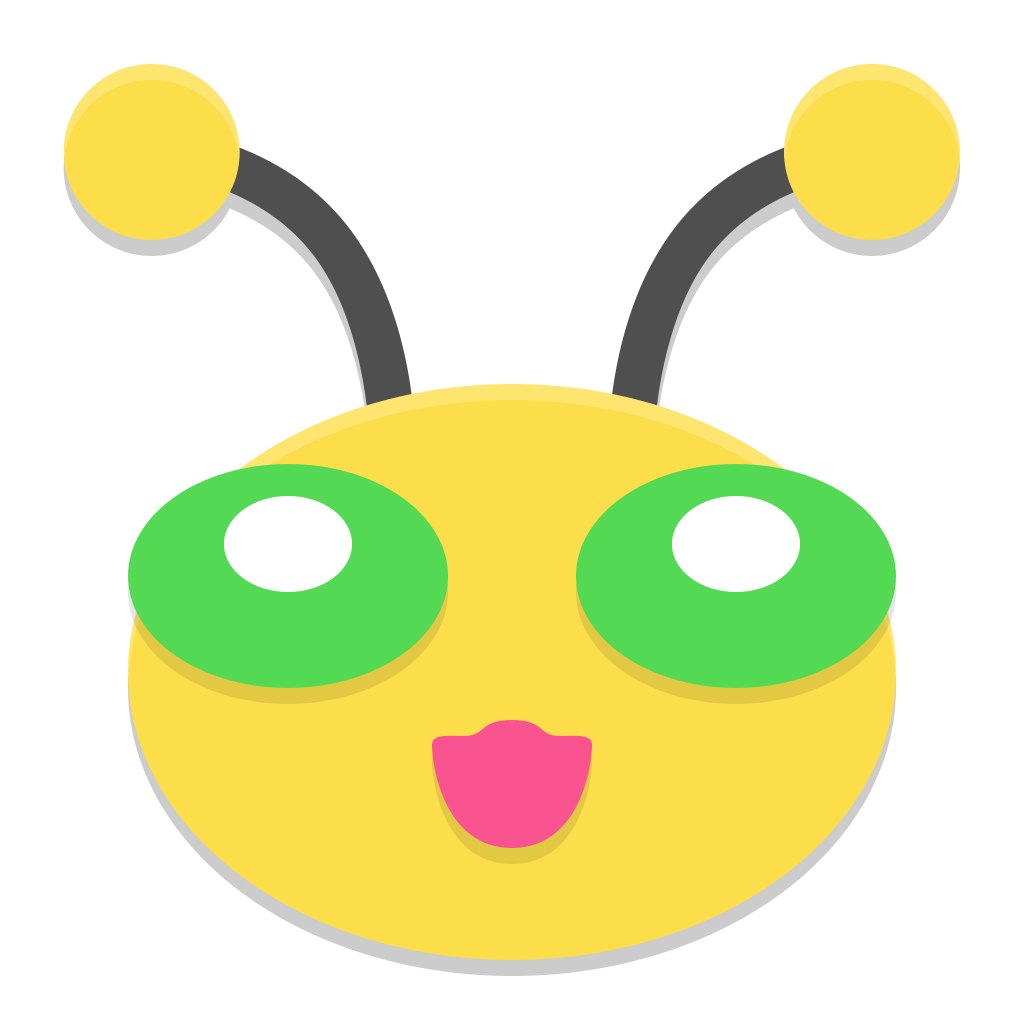} &
\includegraphics[width=\linewidth]{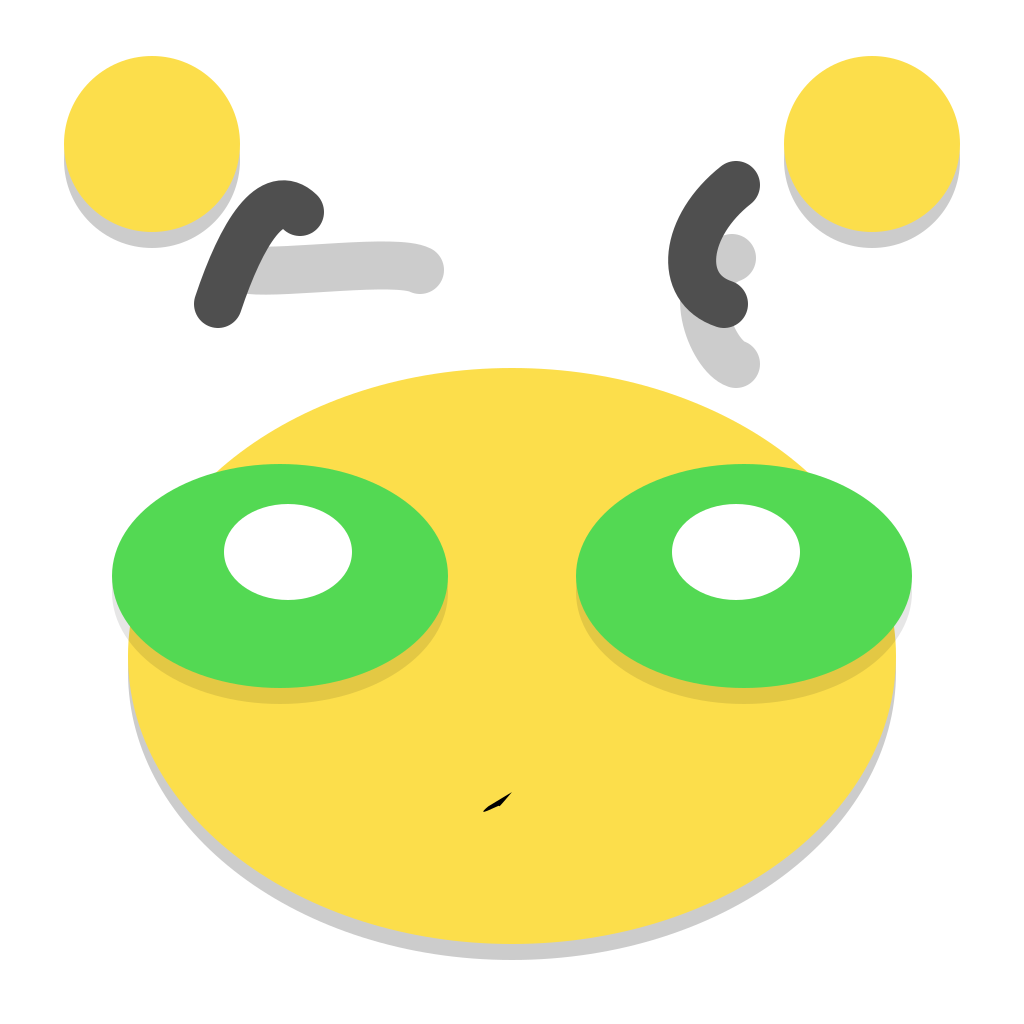} &
\includegraphics[width=\linewidth]{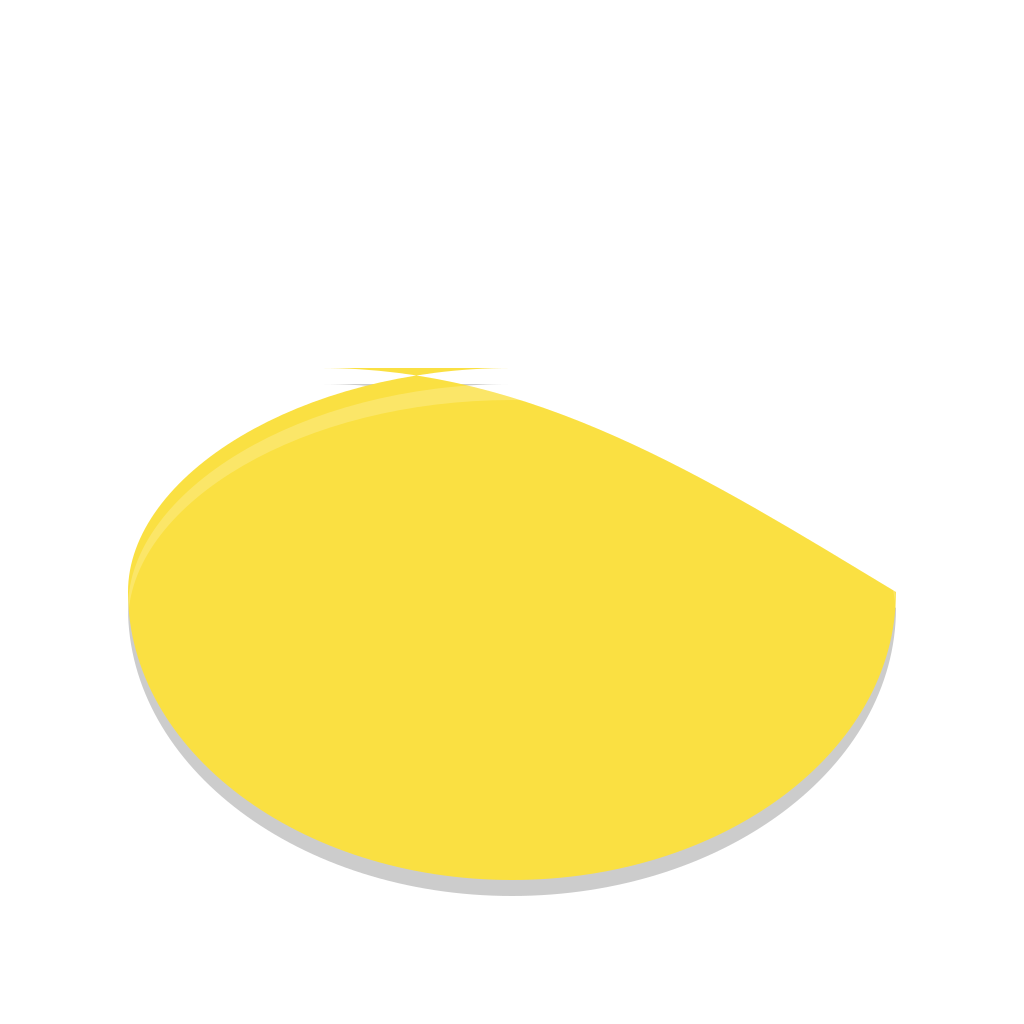} &
\includegraphics[width=\linewidth]{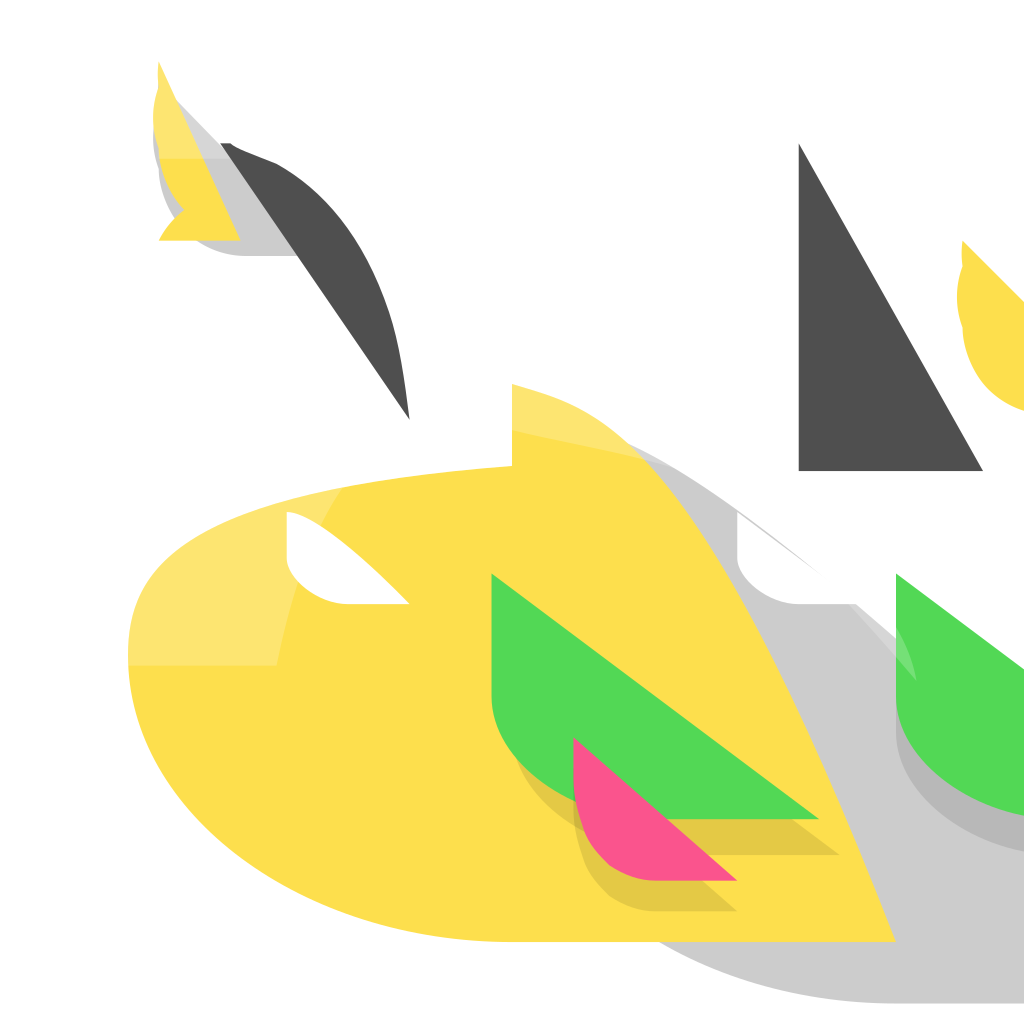} 
\\
\addlinespace[1.5em]

\includegraphics[width=\linewidth]{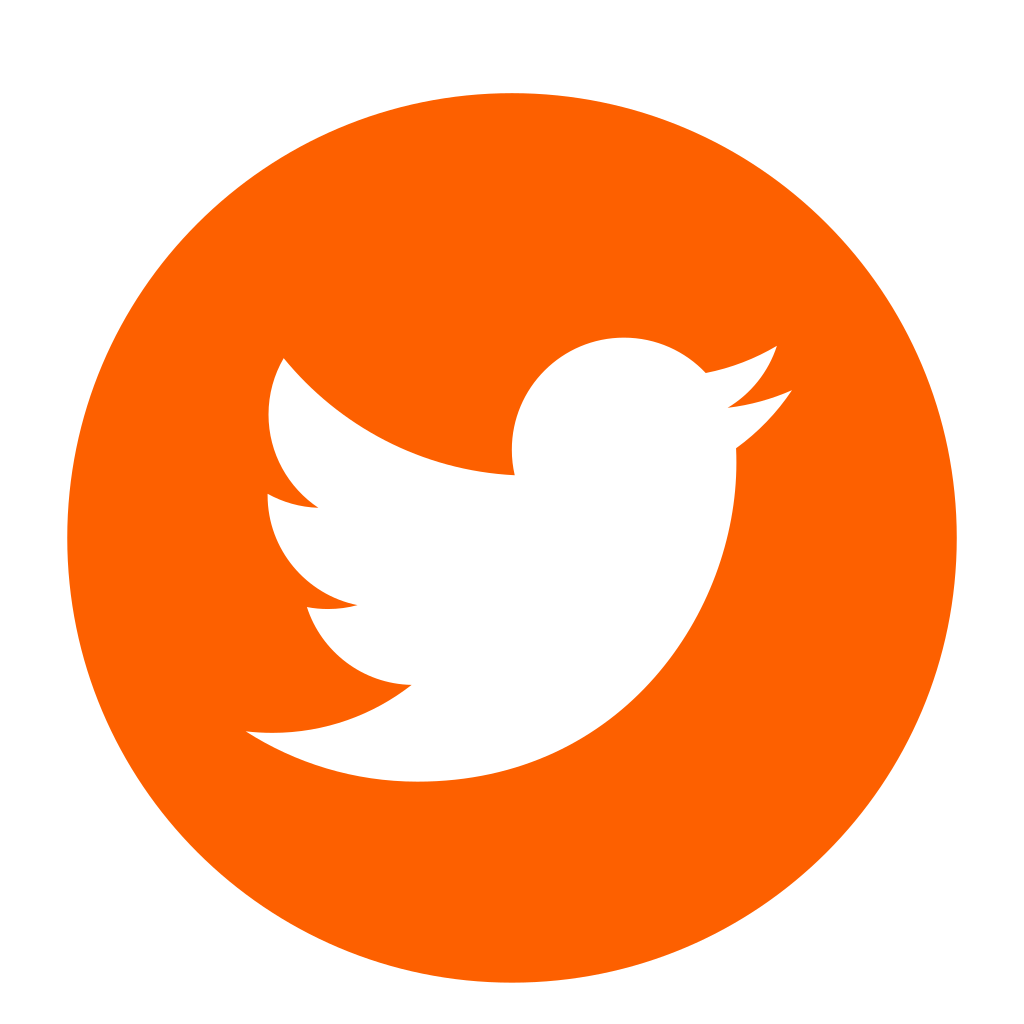} &
\includegraphics[width=\linewidth]{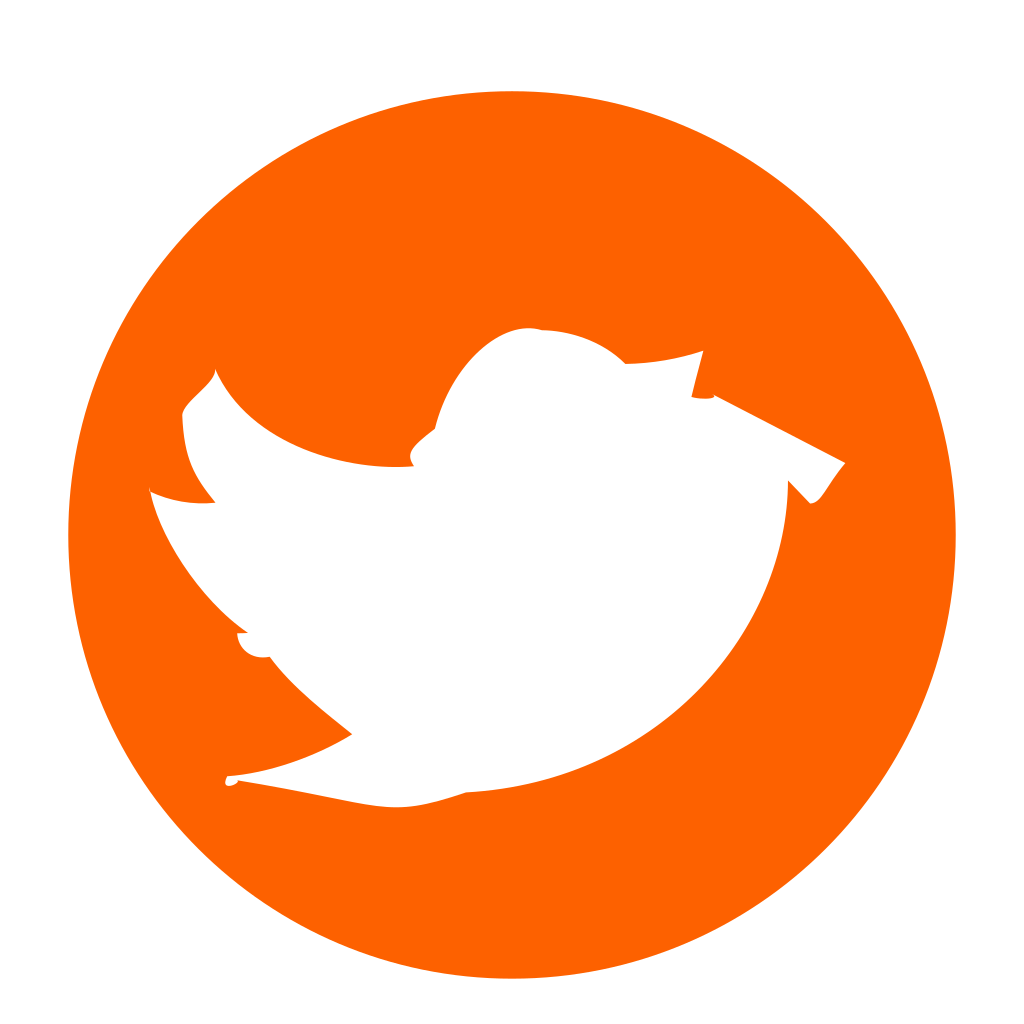} &
\includegraphics[width=\linewidth]{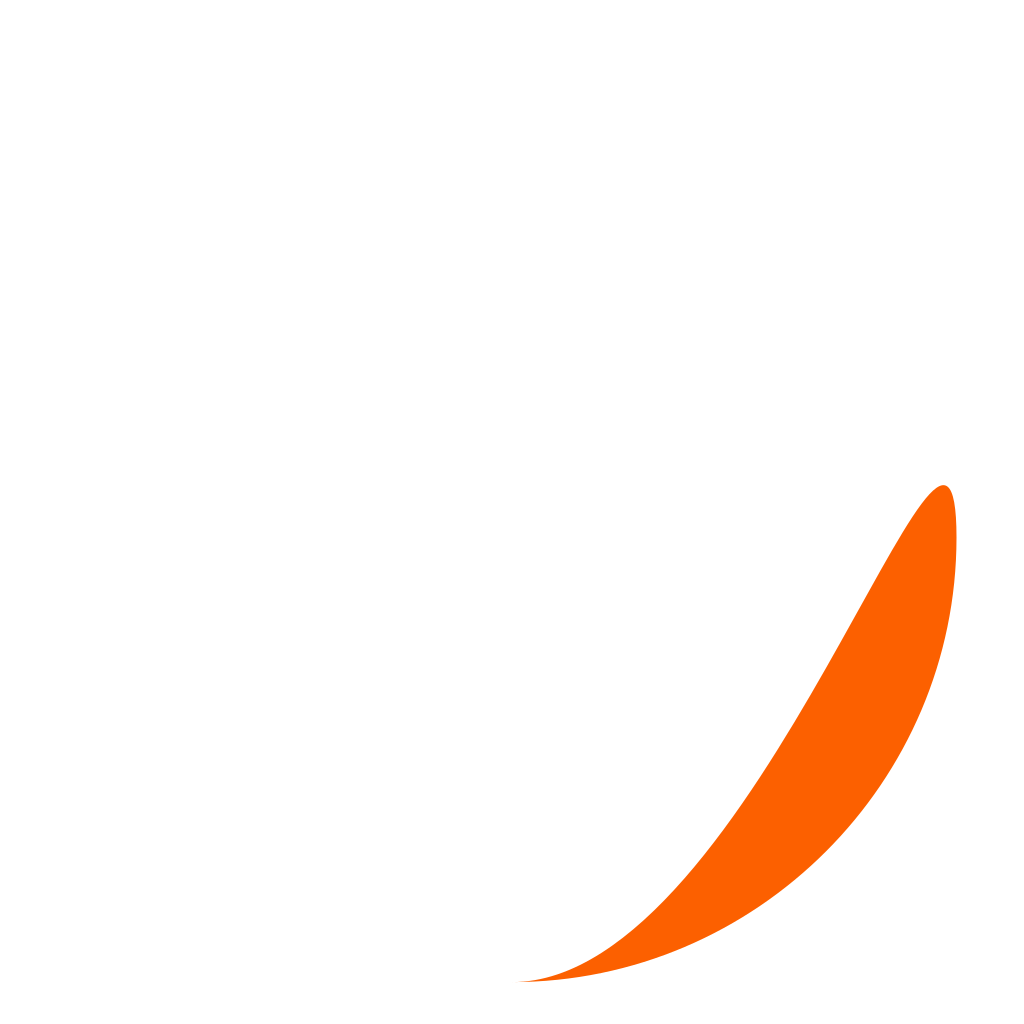} &
\includegraphics[width=\linewidth]{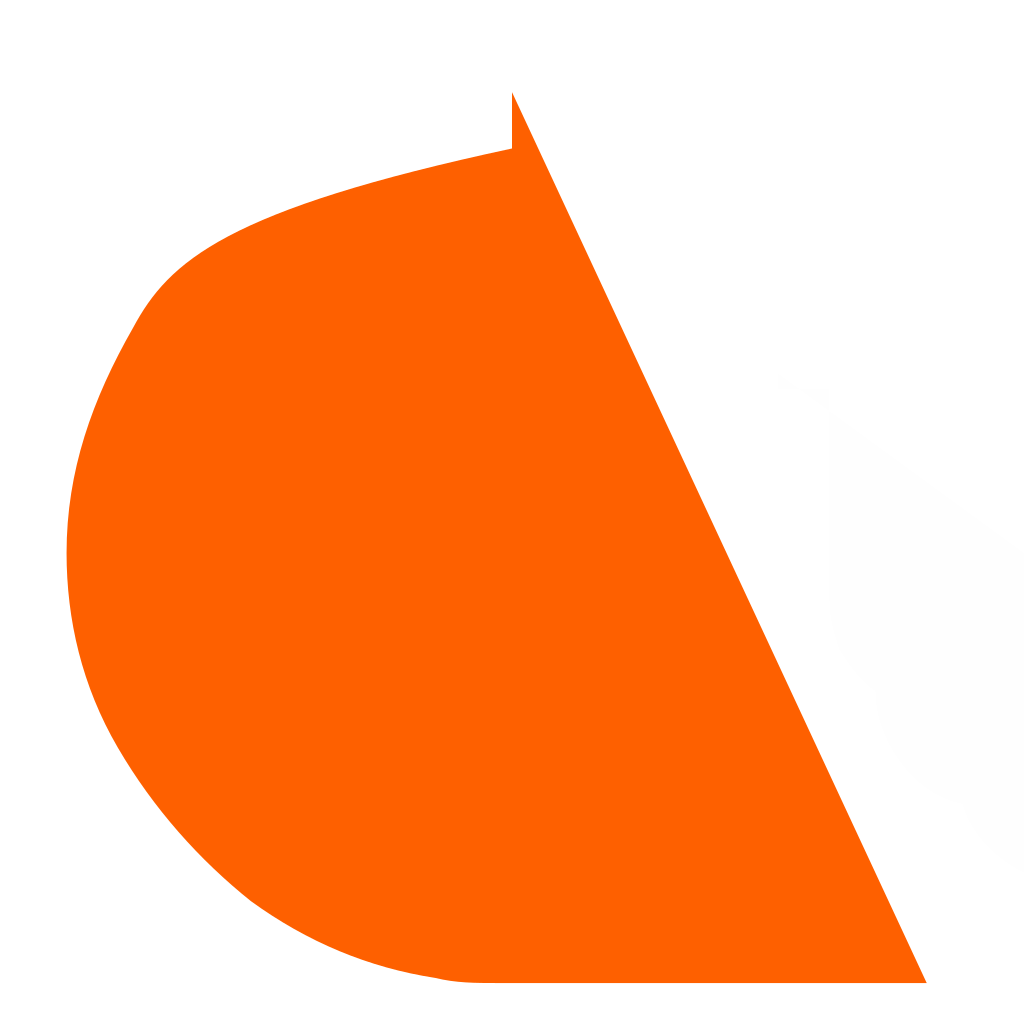} 
\\
\addlinespace[1.5em]

\includegraphics[width=\linewidth]{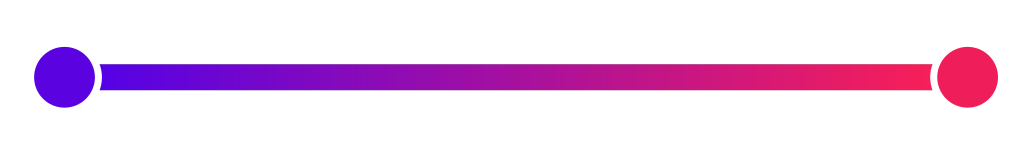} &
\includegraphics[width=\linewidth]{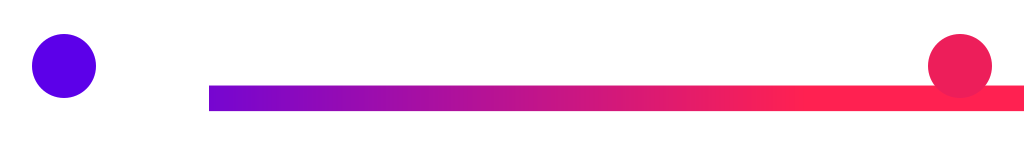} &
\includegraphics[width=\linewidth]{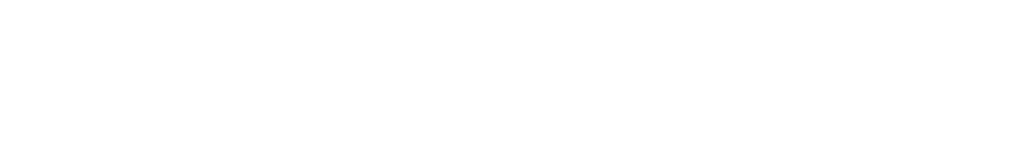} &
\includegraphics[width=\linewidth]{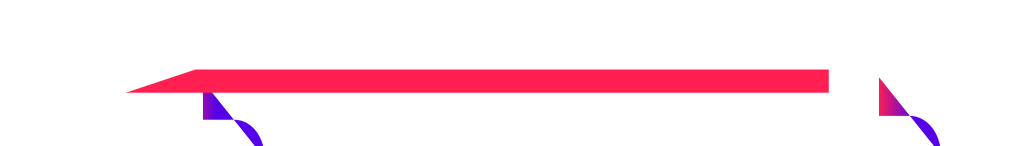} 
\\
\addlinespace[1.5em]

\includegraphics[width=\linewidth]{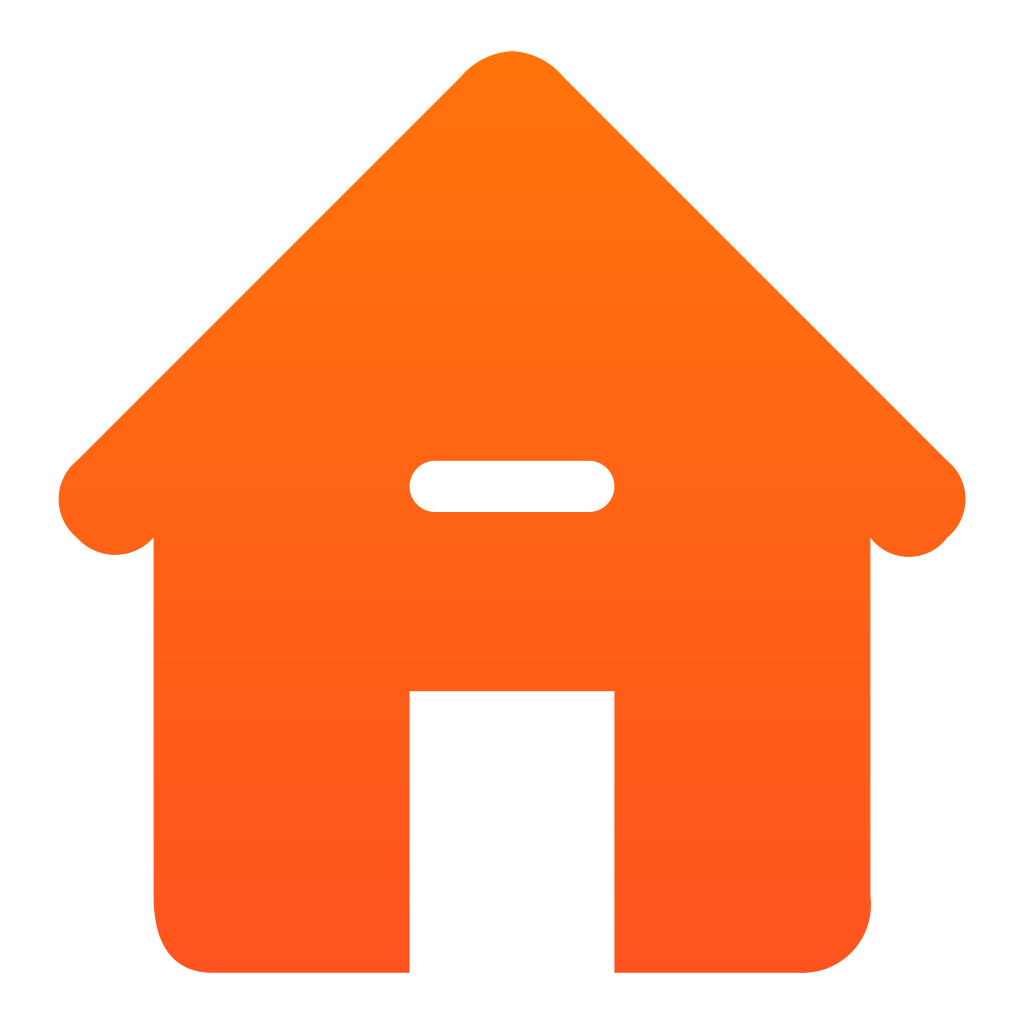} &
\includegraphics[width=\linewidth]{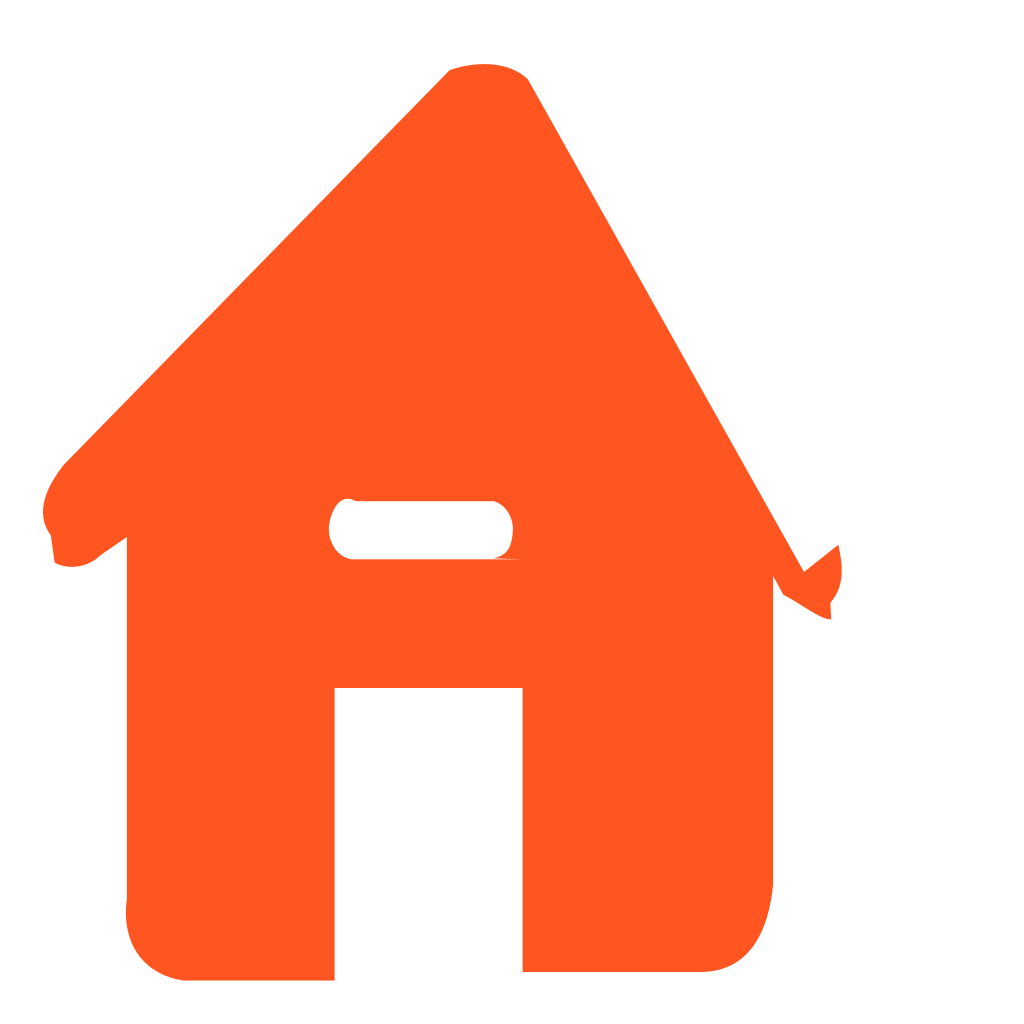} &
\includegraphics[width=\linewidth]{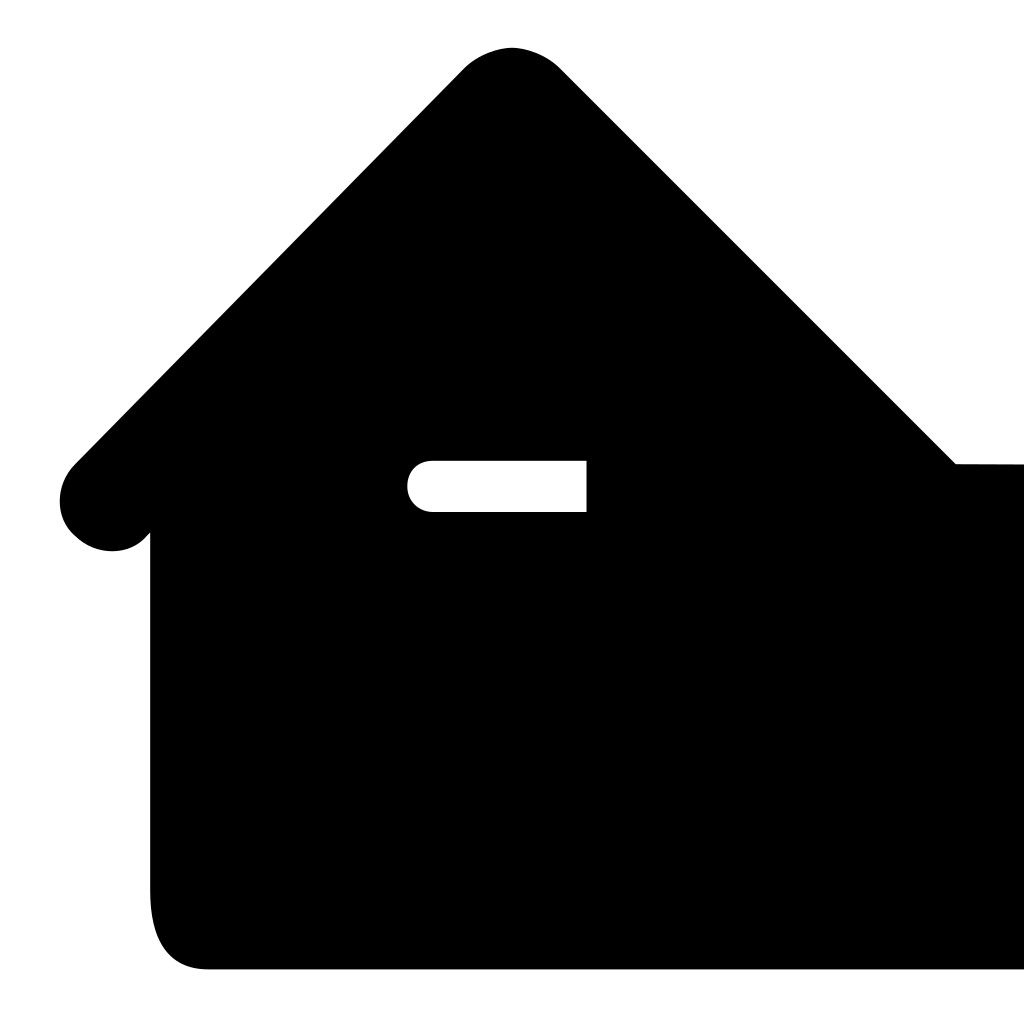} &
\includegraphics[width=\linewidth]{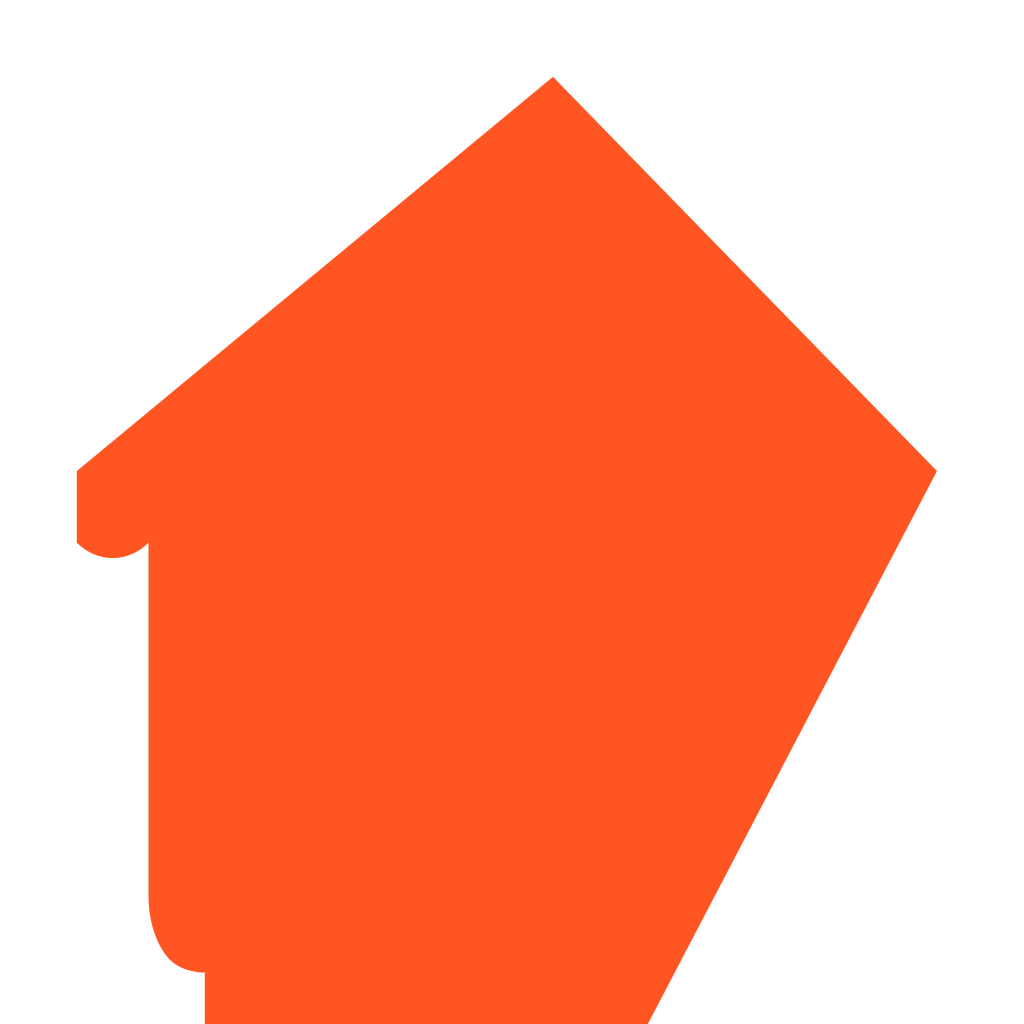} 
\\
\addlinespace[1.5em]

\includegraphics[width=\linewidth]{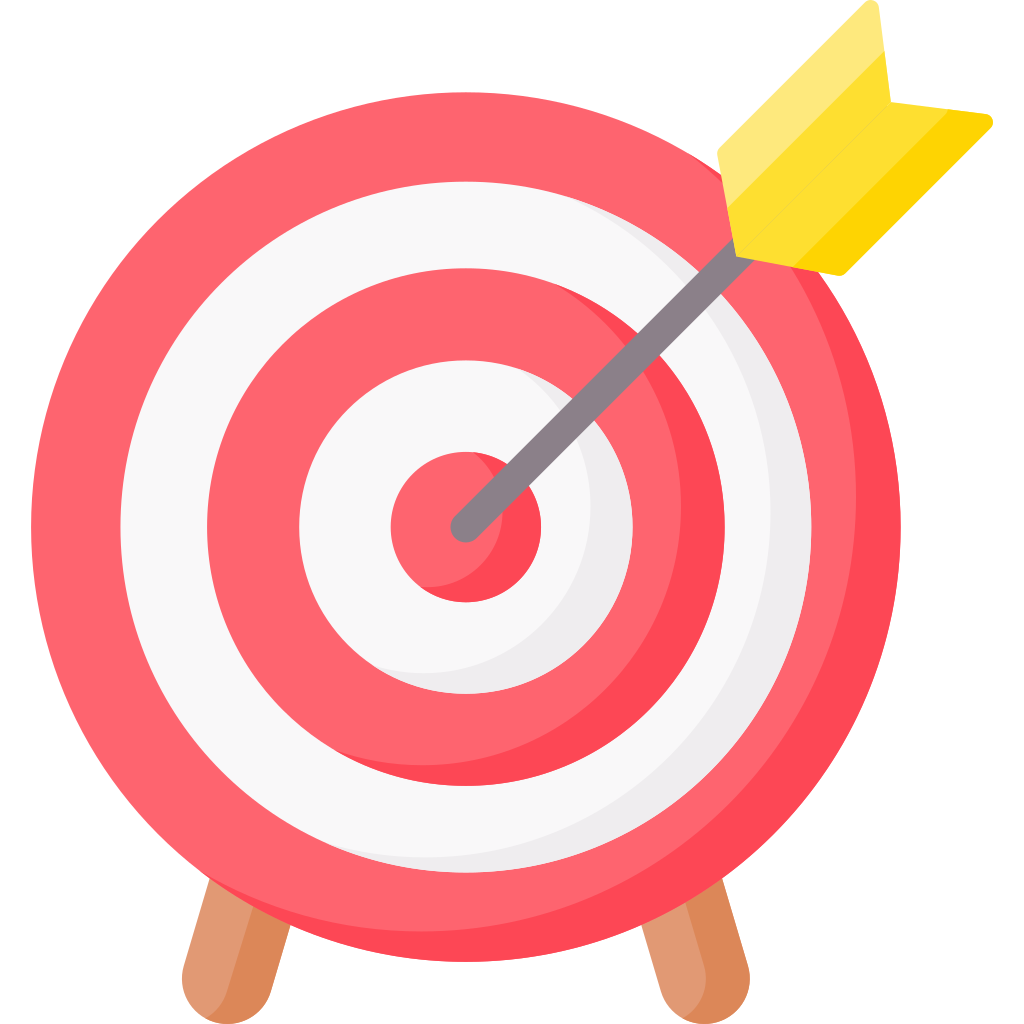} &
\includegraphics[width=\linewidth]{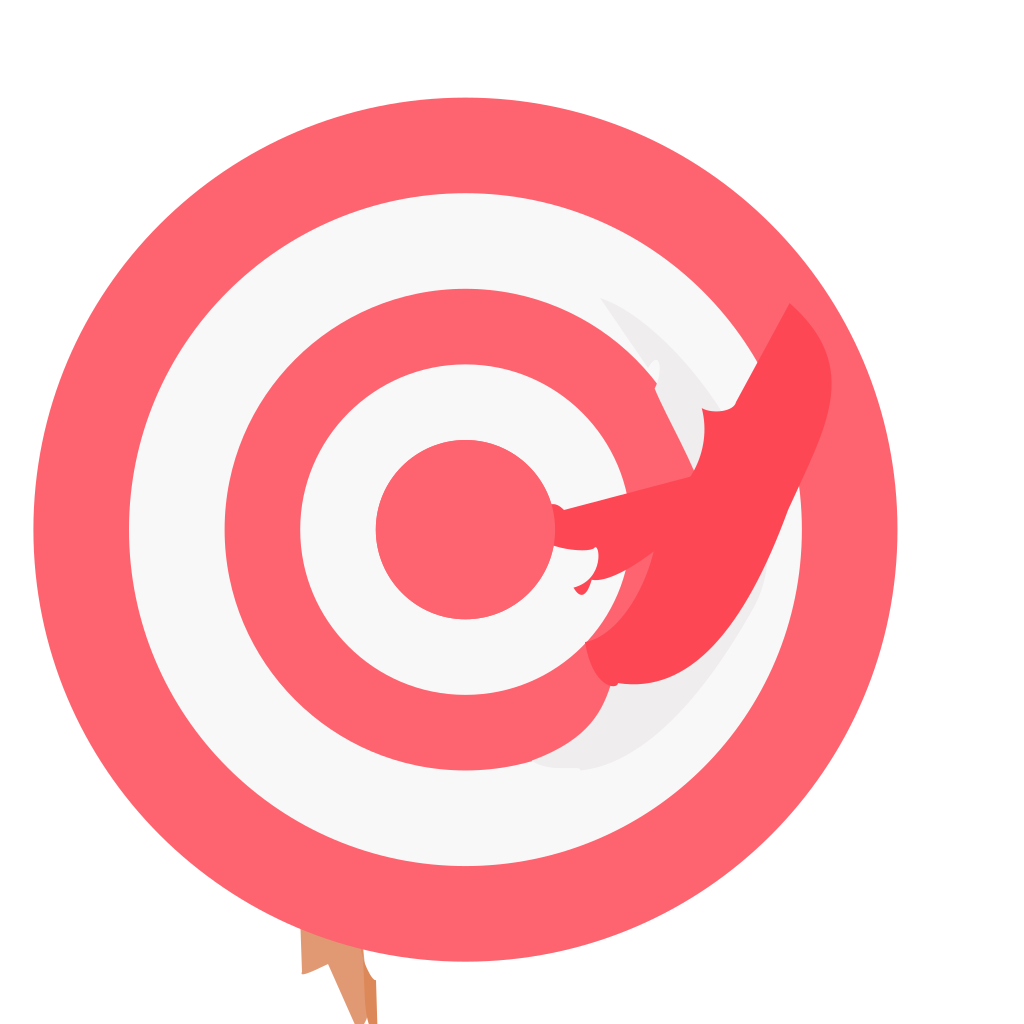} &
\includegraphics[width=\linewidth]{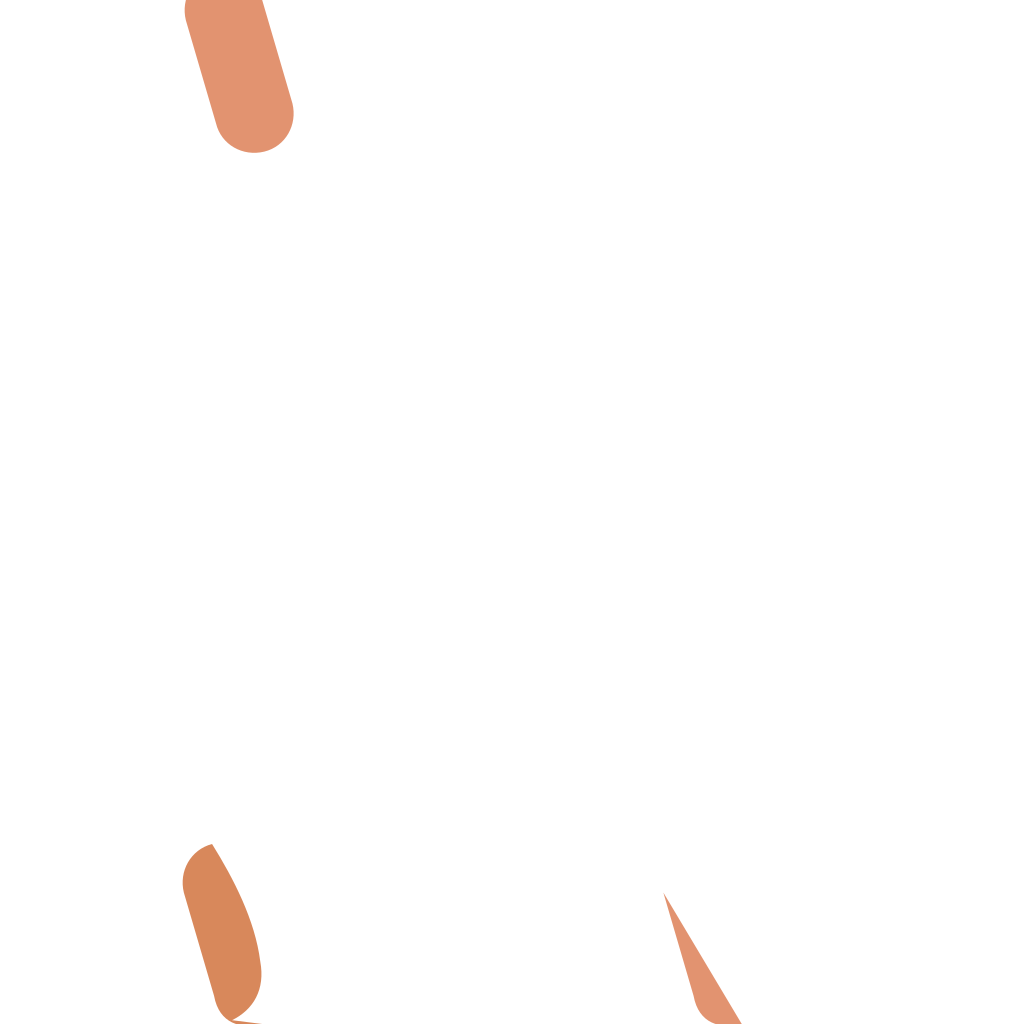} &
\includegraphics[width=\linewidth]{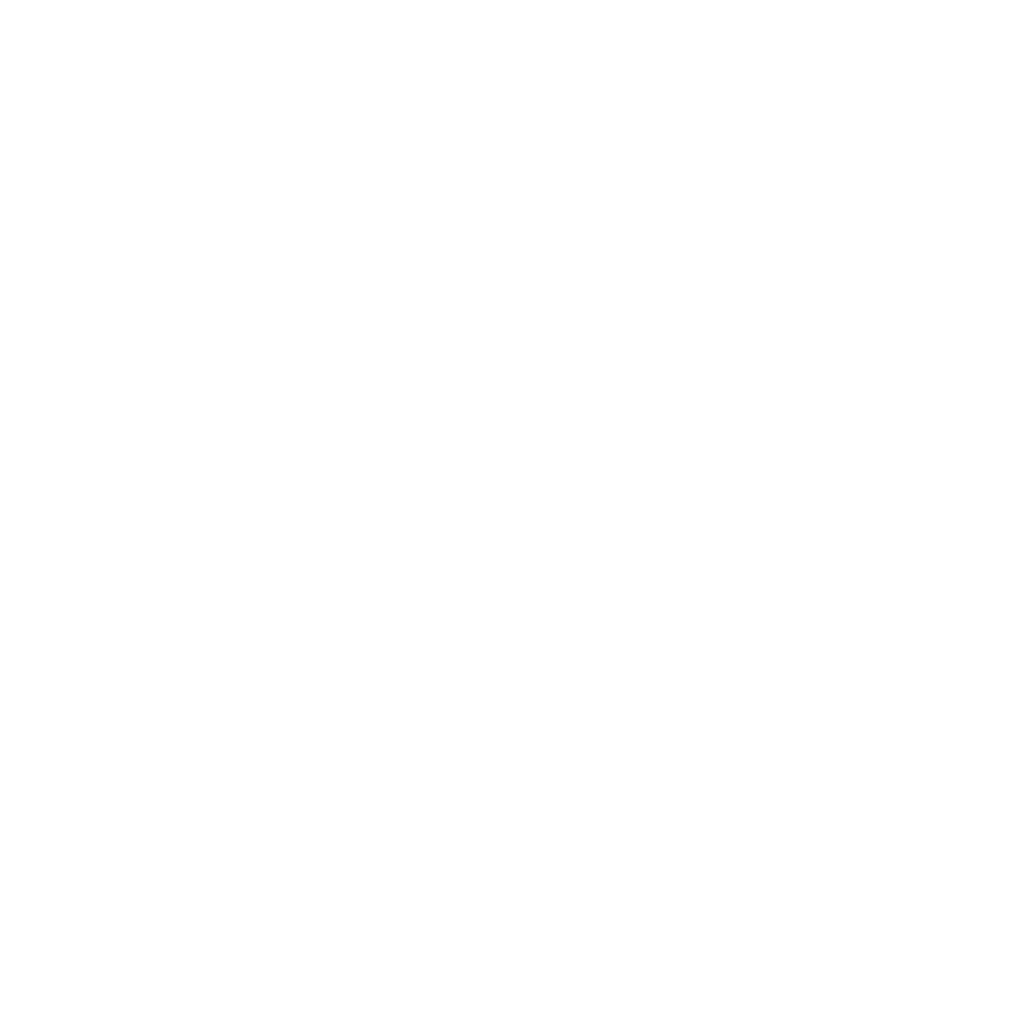} 
\\
\addlinespace[1.5em]

\includegraphics[width=\linewidth]{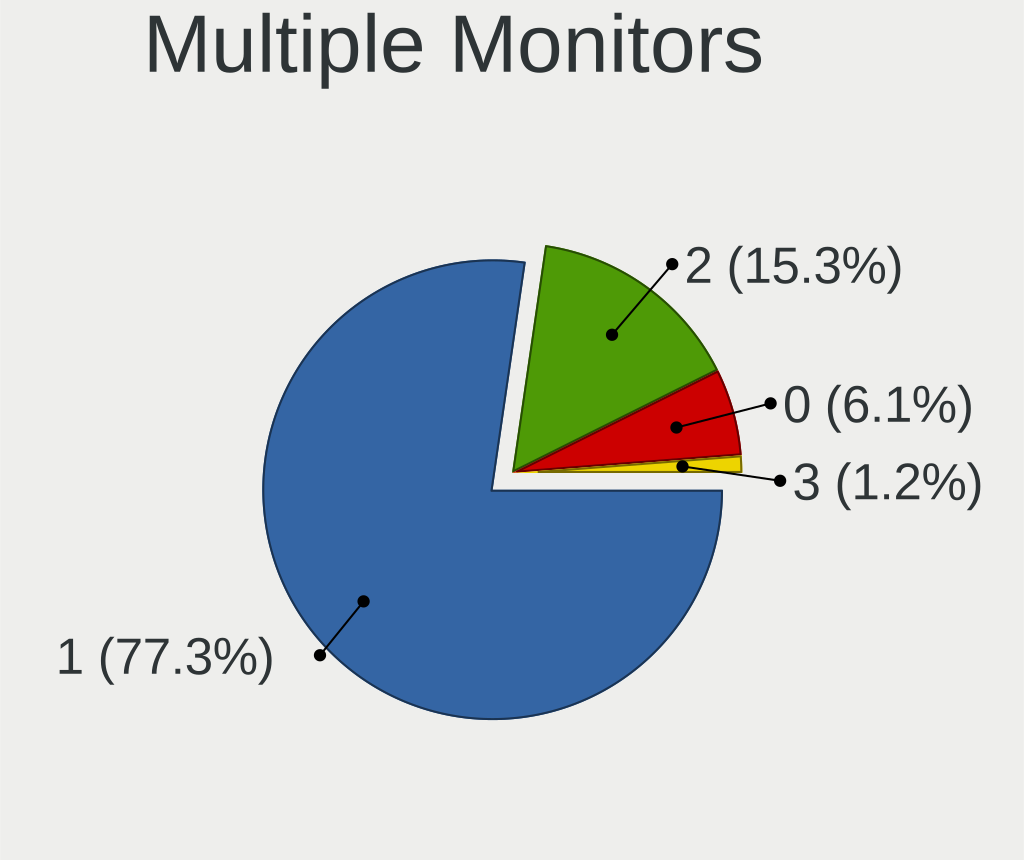} &
\includegraphics[width=\linewidth]{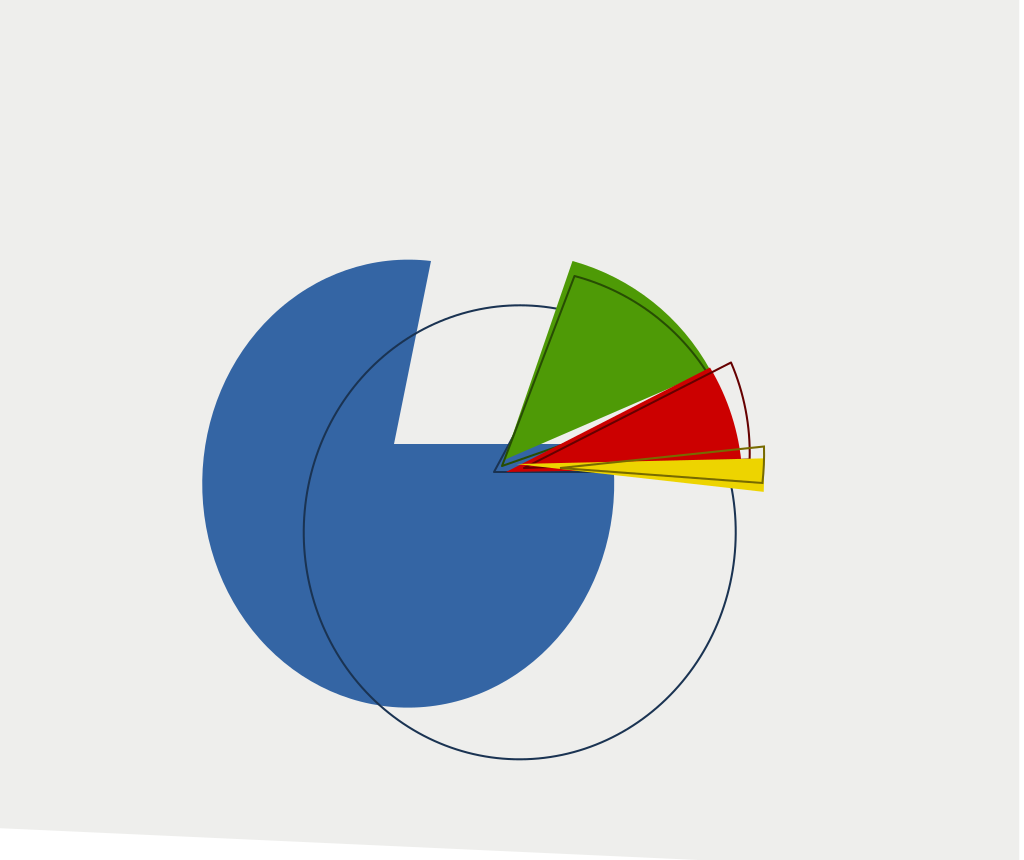} &
\includegraphics[width=\linewidth]{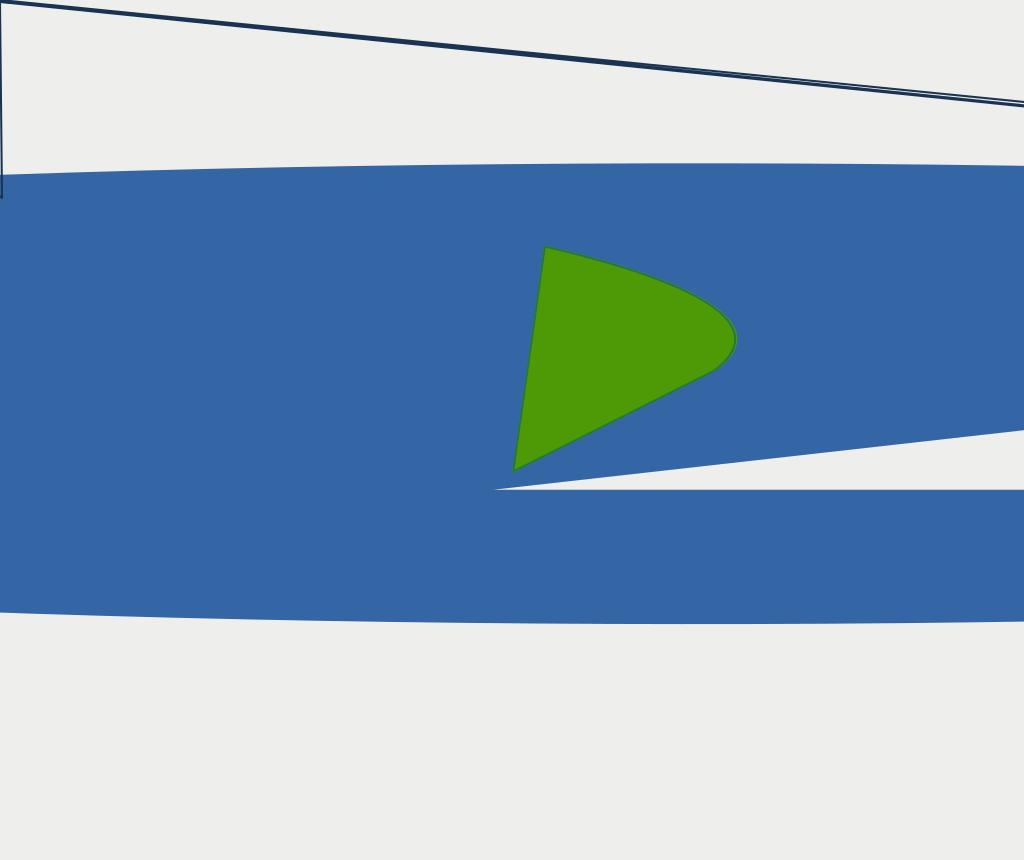} &
\includegraphics[width=\linewidth]{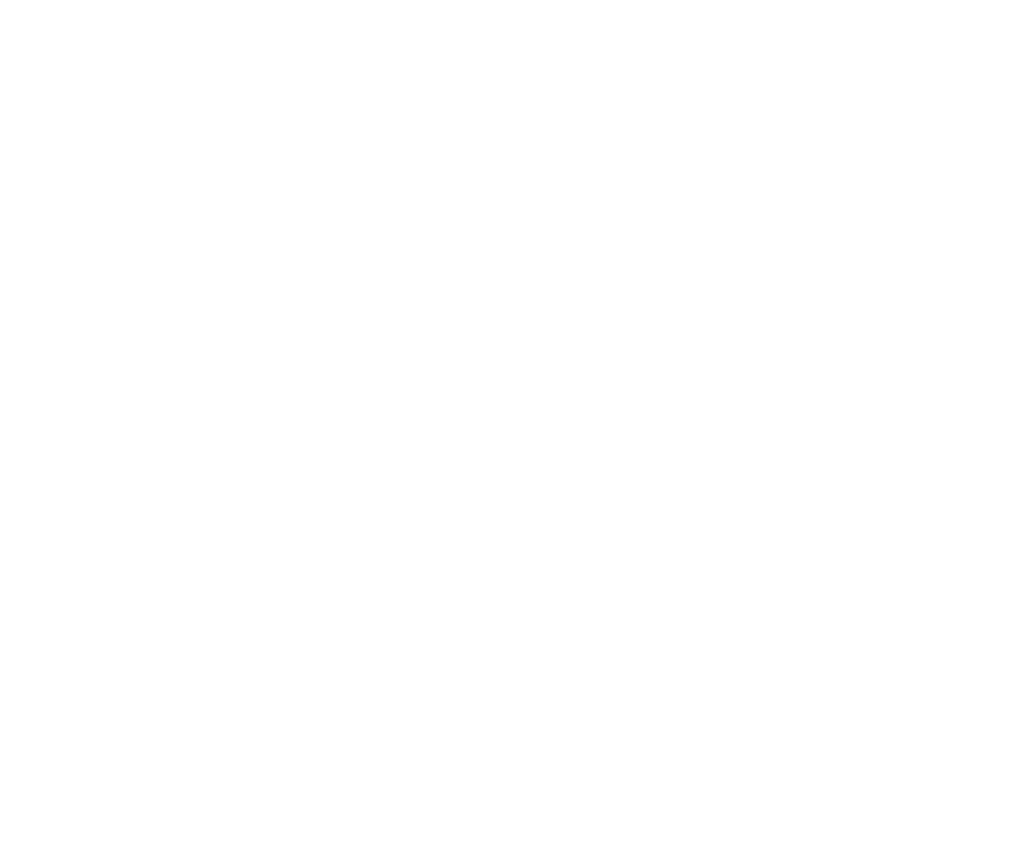} 
\\
\addlinespace[1.5em]

\includegraphics[width=\linewidth]{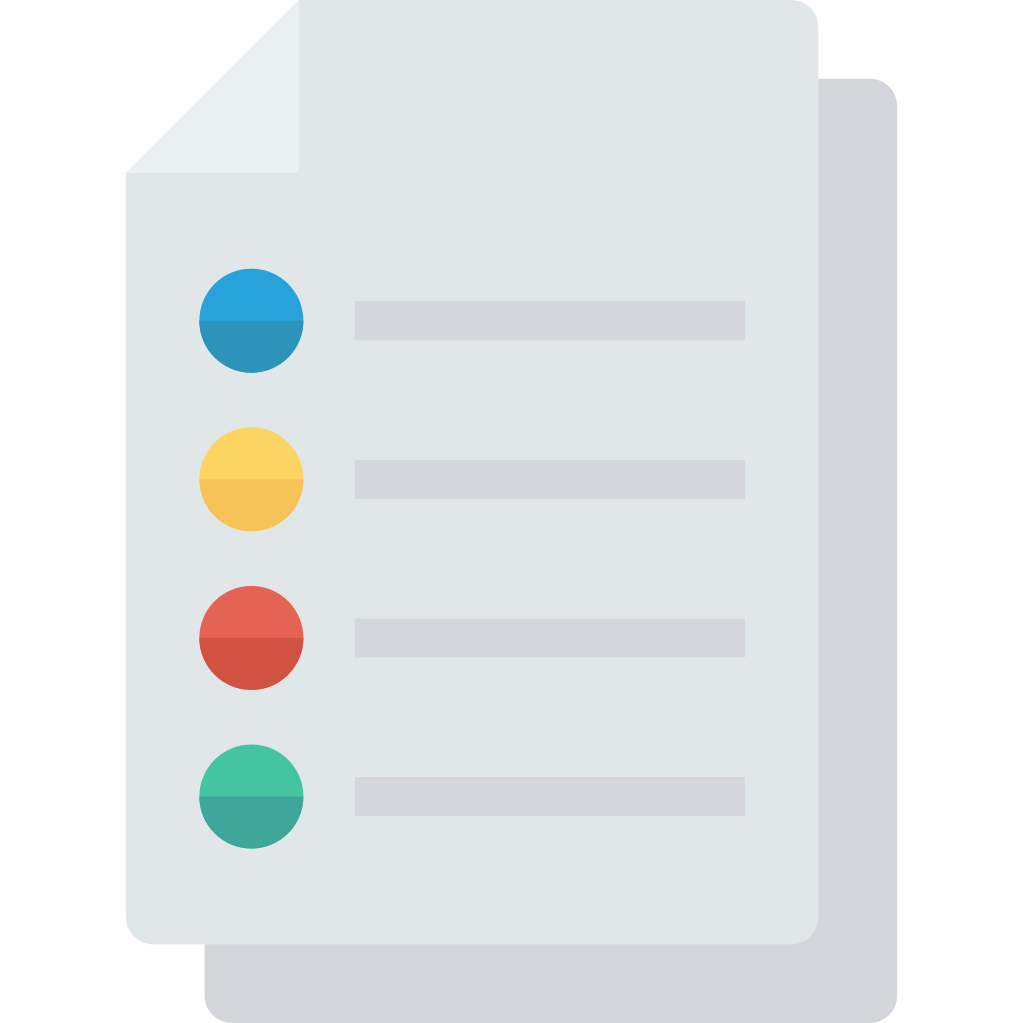} &
\includegraphics[width=\linewidth]{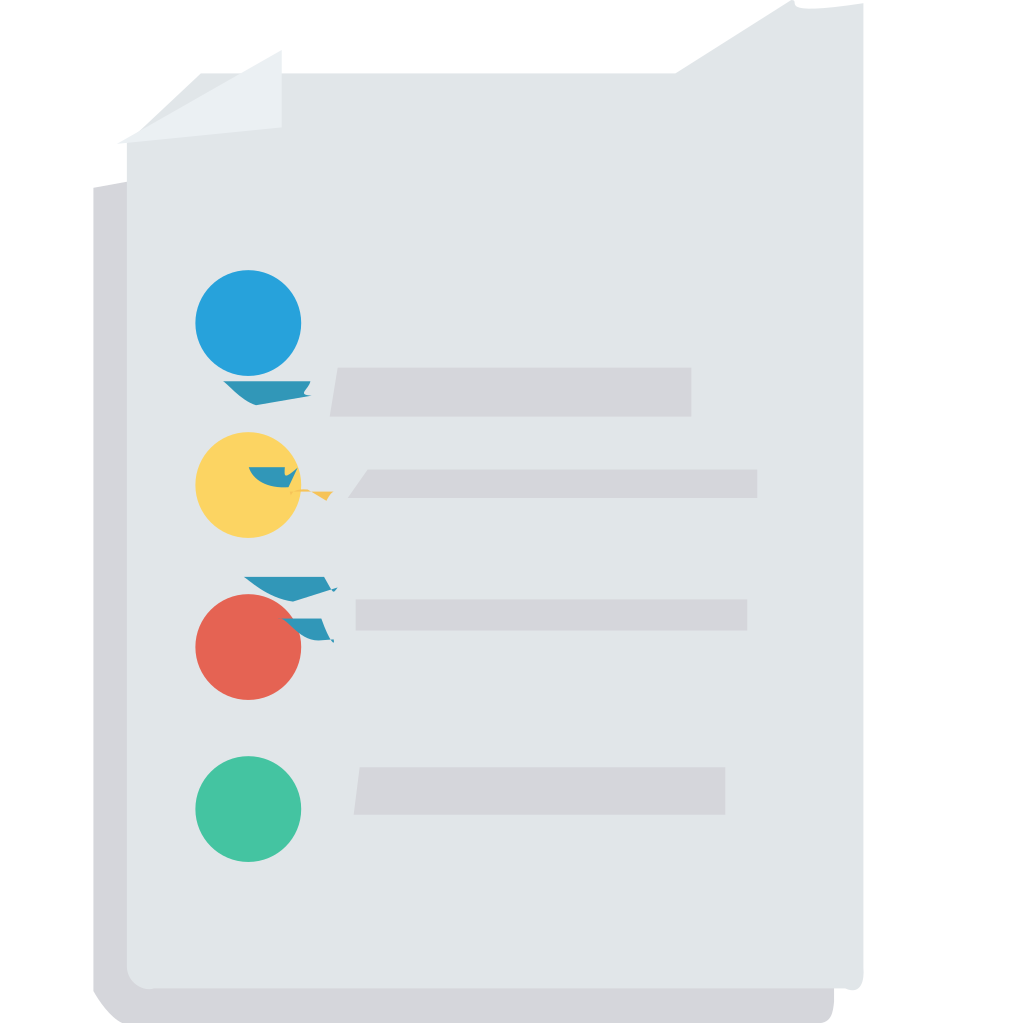} &
\includegraphics[width=\linewidth]{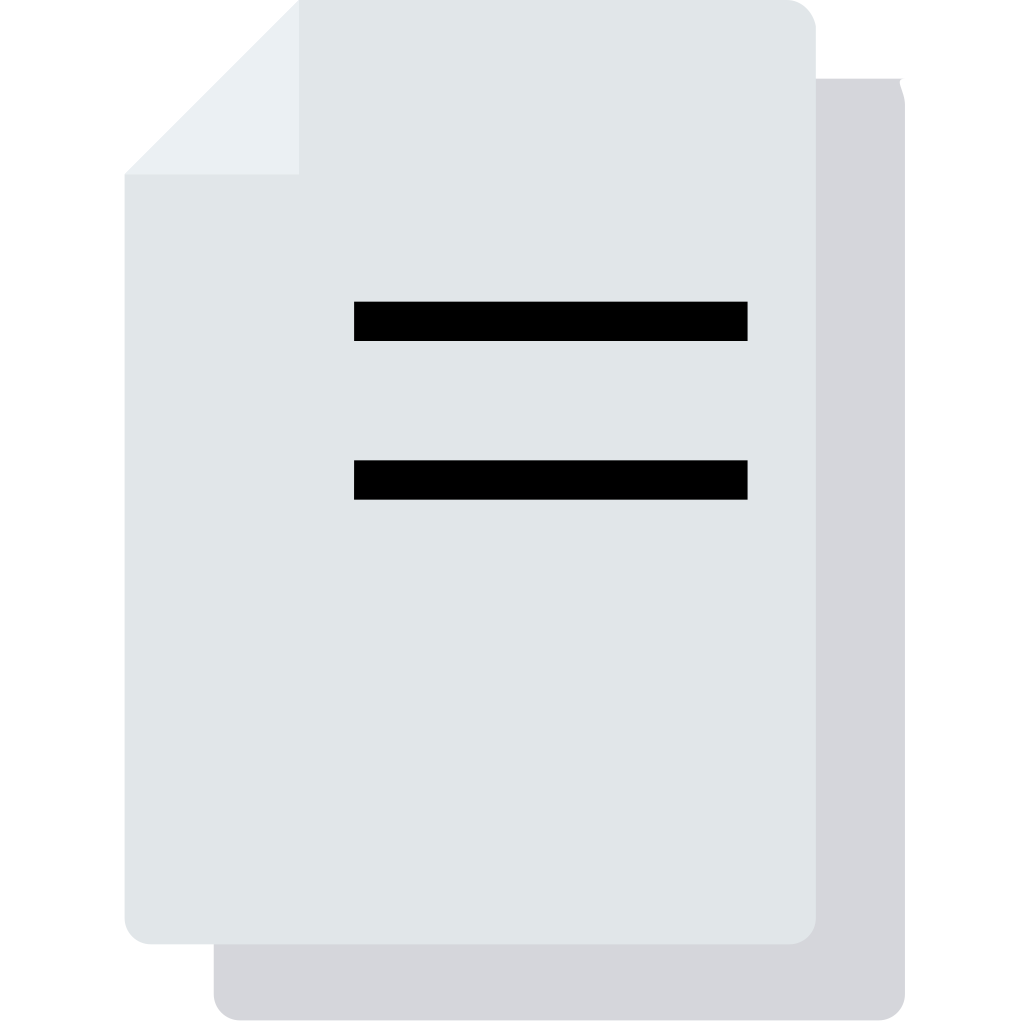} &
\includegraphics[width=\linewidth]{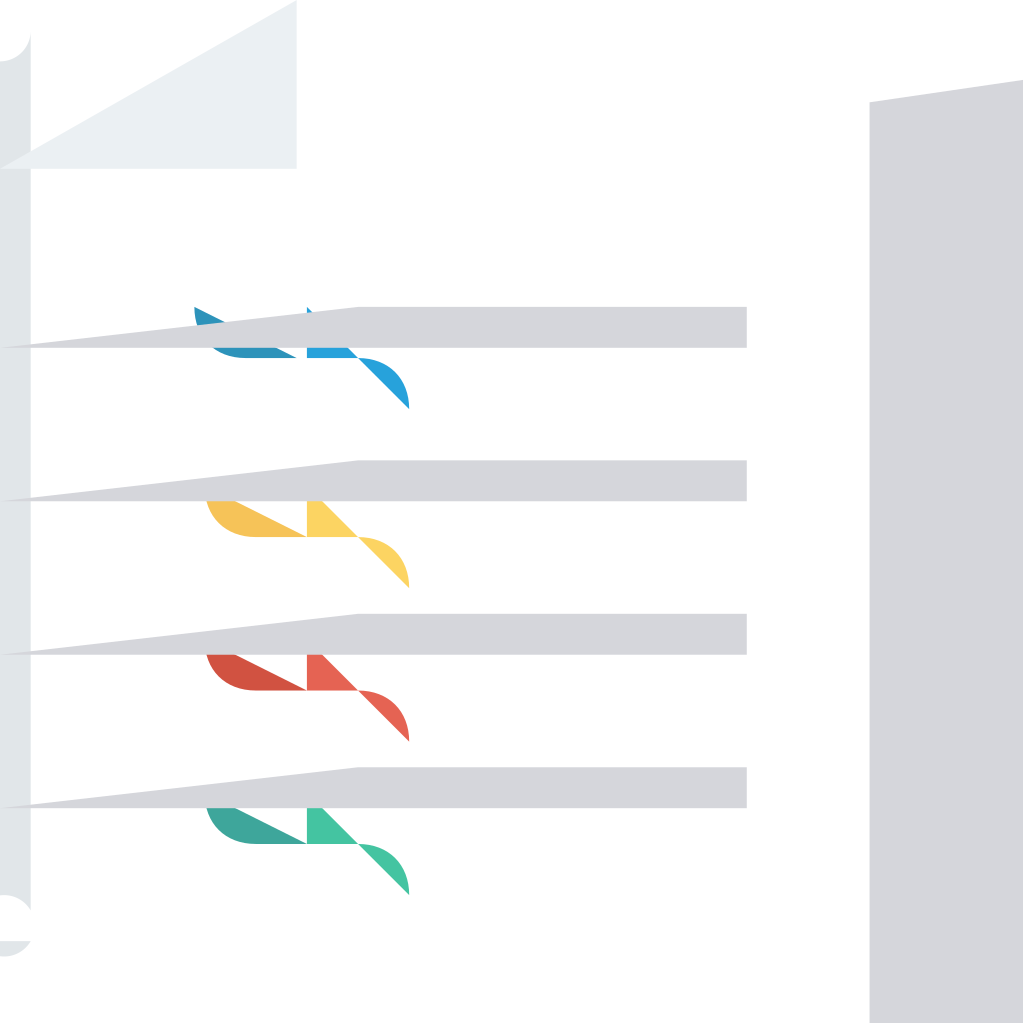} 
\\

\end{longtable}
}

\end{document}